\documentclass{article} 
\PassOptionsToPackage{dvipsnames}{xcolor}
\usepackage[T1]{fontenc}
\usepackage{EMNLP2023}
\usepackage{times}

\usepackage{amsmath,amsfonts,bm}

\def\eqref#1{equation~\ref{#1}}

\def\1{\bm{1}}

\DeclareMathAlphabet{\mathsfit}{\encodingdefault}{\sfdefault}{m}{sl}
\SetMathAlphabet{\mathsfit}{bold}{\encodingdefault}{\sfdefault}{bx}{n}

\usepackage{hyperref}
\usepackage{url}           
\usepackage{booktabs}      
\usepackage{amsfonts}      
\usepackage{nicefrac}       
\usepackage{microtype}      
\usepackage{amsmath}
\usepackage{amssymb}
\usepackage{graphicx}
\usepackage{natbib}
\usepackage{booktabs}
\usepackage{tabularx}
\usepackage{adjustbox}
\usepackage{xspace}
\usepackage{colortbl} 
\usepackage{soul} 
\usepackage{tabularx}
\usepackage{stfloats}
\usepackage{algorithm}
\usepackage[noend]{algpseudocode}
\usepackage{algcompatible}
\usepackage{wrapfig}
\usepackage{pifont}
\usepackage{makecell}
\graphicspath{{./figures/}}
\usepackage{multirow}
\usepackage{xcolor}

\definecolor{lightgrey}{HTML}{dcdbdb}
\sethlcolor{lightgrey}
\definecolor{lightblue}{HTML}{E8F0FE}
\definecolor{lightblue}{HTML}{E8F0FE}
\definecolor{gray}{HTML}{9aa0a6}
\definecolor{lightpink}{HTML}{F48FB1}
\definecolor{lightred}{HTML}{FFCBC9}
\definecolor{lightcyan}{HTML}{80DEEA}
\definecolor{lightgrey}{HTML}{dcdbdb}
\sethlcolor{lightgrey}

\newcommand{\cc}[0]{\cellcolor{lightblue}}

\newcommand{\ModelName}{Multimodal Reasoner as a Judge\xspace}
\newcommand{\ModelNameAbbre}{MR. Judge\xspace}

\newcommand{\Data}{MR-Judge-8K\xspace}
\usepackage{tcolorbox}
\newtcolorbox[auto counter, number within=section, list type=subsubsection, list inside=toc]{sectionbox}[2][]{
colback=white!98!gray, colframe=black, 
colbacktitle=white!90!gray, coltitle=black, 
fonttitle=\bfseries,
title={#2}, 
list entry={Comment \thetcbcounter\quad}
}
\newcommand{\II}{\mathcal{I}}
\newcommand{\RR}{\mathcal{R}}
\newcommand{\LL}{\mathcal{L}}
\newcommand{\TT}{\mathcal{T}}

\title{MR. Judge: Multimodal Reasoner as a Judge}

\author{Renjie Pi$^1$, \quad Felix Bai$^2$, \quad Qibin Chen$^2$, \quad Simon Wang$^2$, \quad Jiulong Shan$^2$\\
\quad\textbf{Kieran Liu}$^2$, \quad \textbf{Meng Cao$^2$}
\\
  $^1$HKUST\quad
$^2$Apple}

\begin{document}

\maketitle
\begin{abstract} The paradigm of using Large Language Models (LLMs) and Multimodal Large Language Models (MLLMs) as evaluative judges has emerged as an effective approach in RLHF and inference-time scaling. In this work, we propose \textbf{\ModelName} (\ModelNameAbbre), a paradigm for empowering general-purpose MLLMs judges with strong reasoning capabilities. Instead of directly assigning scores for each response, we formulate the judgement process as a reasoning-inspired multiple-choice problem. Specifically, the judge model first conducts deliberate reasoning covering different aspects of the responses and eventually selects the best response from them. This reasoning process not only improves the interpretibility of the judgement, but also  greatly enhances the performance of MLLM judges. To cope with the lack of questions with scored responses, we propose the following strategy to achieve automatic annotation: 1) Reverse Response Candidates Synthesis: starting from a supervised fine-tuning (SFT) dataset, we treat the original response as the best candidate and prompt the MLLM to generate plausible but flawed negative candidates. 
2) Text-based reasoning extraction: we carefully design a data synthesis pipeline for distilling the reasoning capability from a text-based reasoning model, which is adopted to enable the MLLM judges to regain complex reasoning ability via warm up supervised fine-tuning. Experiments demonstrate that our \ModelNameAbbre is effective across a wide range of tasks. Specifically, our \ModelNameAbbre-7B surpasses GPT-4o by 9.9\% on VL-RewardBench, and improves performance on MM-Vet during inference-time scaling by up to 7.7\%.

\end{abstract}
\begin{figure}[t!]
\includegraphics[width=.48\textwidth]{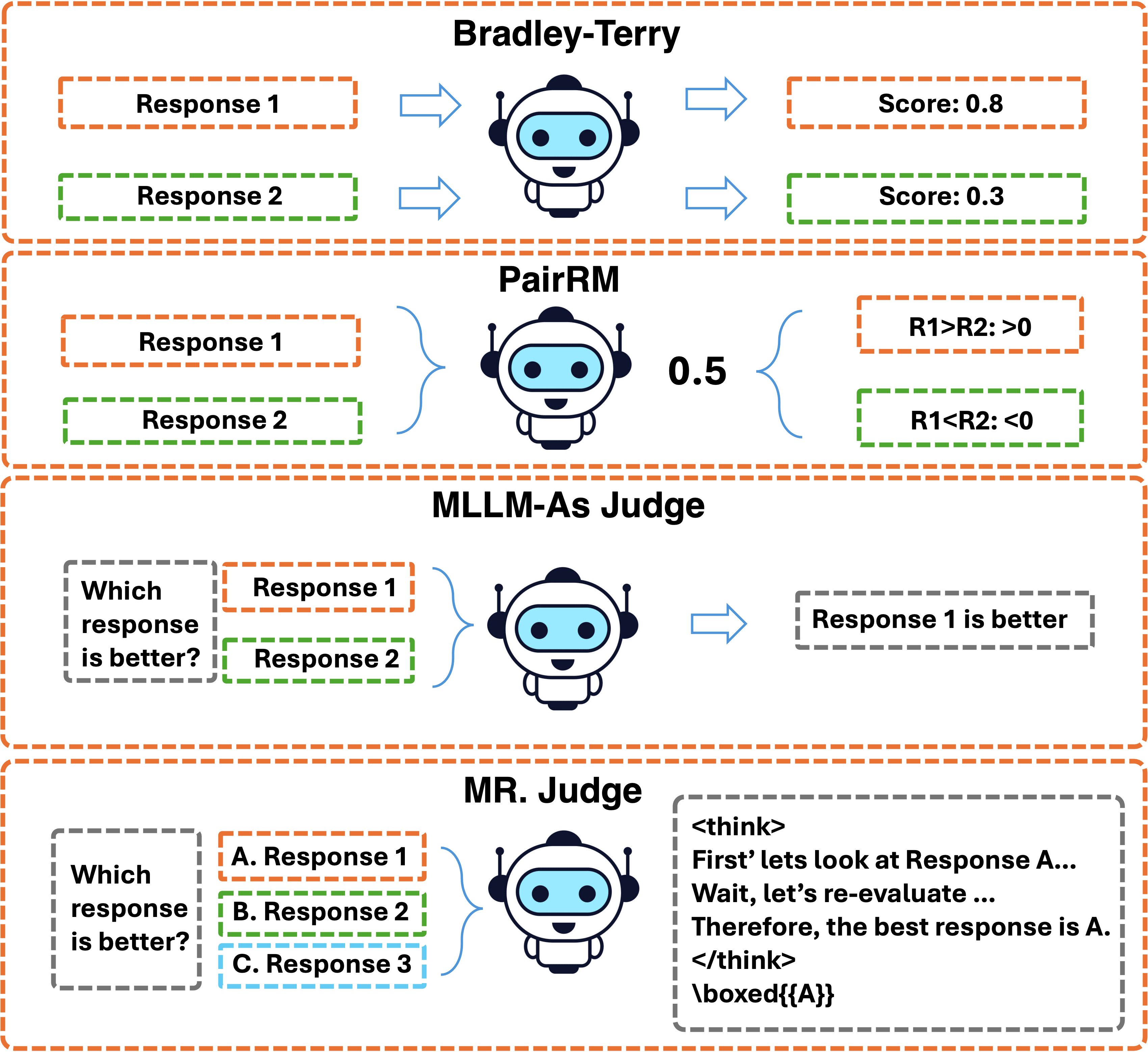} 
\vspace{-0.6cm}
\caption{ Illustration of different reward models and judge models.
}\label{fig:teaser}
\vspace{-1.5em}
\end{figure}

\begin{figure*}[h]
\includegraphics[width=1.0\textwidth]{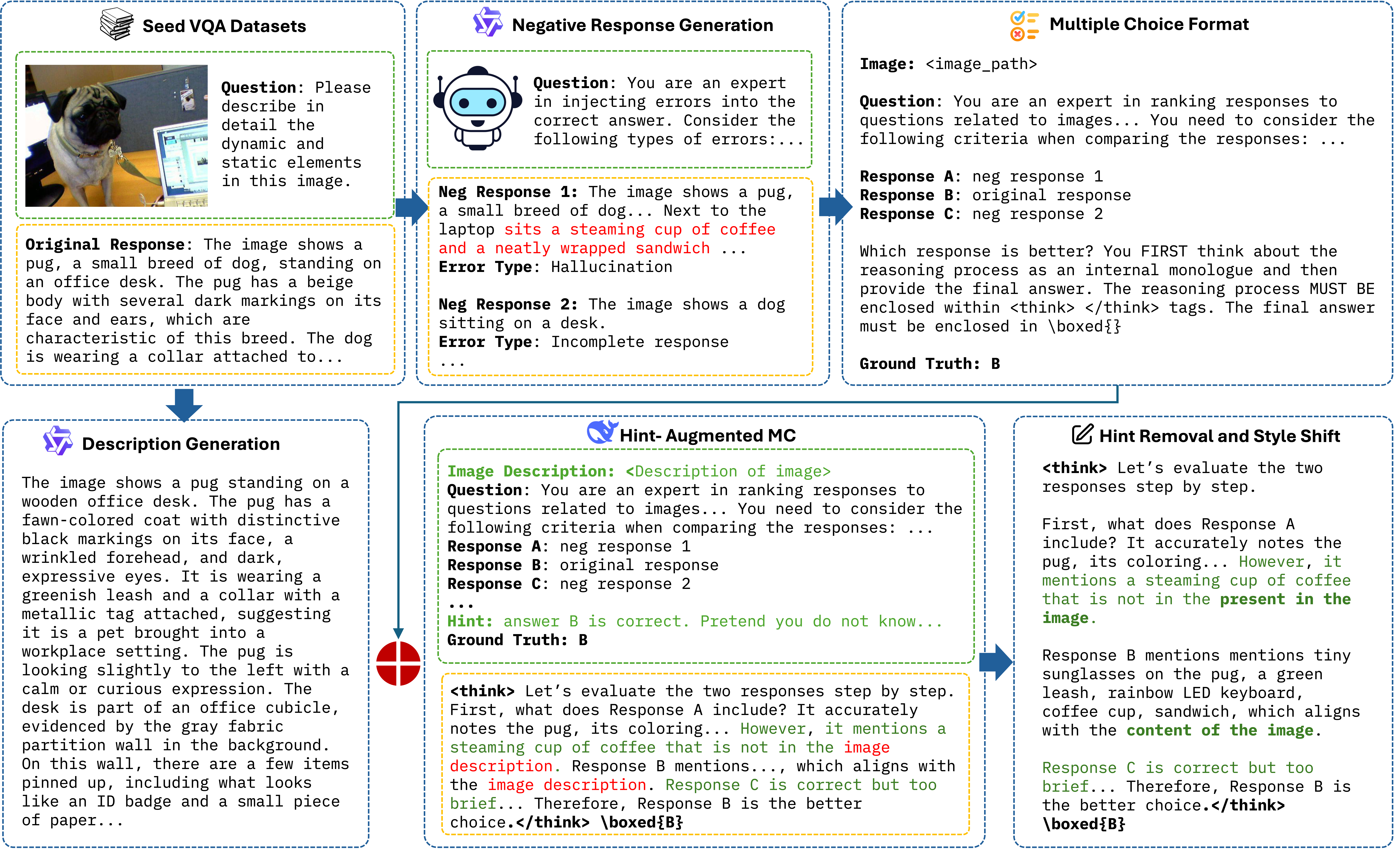} 
\vspace{-0.6cm}
\caption{ Overview of the data generation framework. The top row illustrates the synthesis of negative response candidates using the MLLM, followed by the construction of multiple-choice questions from the image, the prompt, and the candidate responses. The bottom row depicts how long-form reasoning data is extracted from text-based reasoning LLMs for supervised fine-tuning.
}\label{fig:framework}
\vspace{-1.5em}
\end{figure*}

\section{Introduction}
Effective reward modeling is particularly valuable in reinforcement learning (RL), where large language model (LLM) judges can provide real-time feedback or rank responses during reinforcement learning from human feedback (RLHF), thereby accelerating policy optimization~\citep{yuan2025selfrewardinglanguagemodels}. It is also beneficial for inference-time scaling, where LLM judges enable multimodal LLMs (MLLMs) to perform multiple inference passes and select the best output among candidate responses~\citep{xie2023selfevaluationguidedbeamsearch}. Traditional score-based reward models typically append a classifier to the LLM output, predicting a single scalar score to represent response quality~\citep{liu2024skyworkrewardbagtricksreward, li2023reinforcementlearninghumanfeedback}. While this approach has proven effective, the resulting scores are often not interpretable—failing to explain why certain responses receive high or low ratings—which introduces risks such as reward hacking~\citep{skalse2025definingcharacterizingrewardhacking}. In contrast, using LLMs as judges allows for more interpretable and diverse evaluation processes~\citep{xie2023selfevaluationguidedbeamsearch, yao2023treethoughtsdeliberateproblem, liu2025inferencetimescalinggeneralistreward, hosseini2024vstartrainingverifiersselftaught}. These models not only select the best response but also provide reasoning behind their decisions, offering greater transparency and robustness in evaluation.

On the other hand, the use of multimodal large language models (MLLMs) as judges~\citep{chen2024mllmasajudgeassessingmultimodalllmasajudge} still presents several challenges. First, instruction-tuned MLLMs are generally less capable of evaluating the quality of model-generated responses. This limitation stems from the significantly smaller scale of datasets used during multimodal training compared to those used for text-only LLMs, resulting in weaker multimodal capabilities~\citep{chen2024quantifyingmitigatingunimodalbiases, pi2024strengtheningmultimodallargelanguage}. Second, due to the unreliable capability of SOTA MLLMs to score responses, it is challenging to scale up the preference dataset with high quality by leveraging MLLMs as annotators, making it difficult to train effective judge models. Third, instruction-tuned MLLMs often struggle to produce complex reasoning traces, which are critical for assessing responses to more challenging or nuanced questions~\citep{xu2025llavacotletvisionlanguage}.

In this work, we propose \ModelName (\ModelNameAbbre), a novel approach for training general-purpose judges to evaluate responses from MLLMs, particularly suited for complex, open-ended tasks where exact ground-truth answers are unavailable or inherently ambiguous. Our method adopts a generative reasoning paradigm in which the judge model not only selects the best response but also engages in detailed reasoning over multiple candidate responses prior to making a decision. As illustrated in Figure~\ref{fig:teaser}, we formulate the judgment task as a multiple-choice selection problem associated with explicit reasoning, which offers several key advantages: 1) the multiple-choice structure encourages comparative reasoning between multiple candidate responses, which has been shown to enhance judgment accuracy in preference-based tasks~\citep{jiang2023llmblenderensemblinglargelanguage}; 2) The detailed reasoning traces prior to the final decision help mitigate problems such as reward hacking (e.g., prefer lengthy responses), since the reasoning allows the judgment depend on critical aspects of the responses, rather than some spurious features; 3) This formulation enables the use of rule-based supervision signals—similar to DeepSeek-R1~\citep{deepseekai2025deepseekr1incentivizingreasoningcapability, zeng2025simplerlzooinvestigatingtamingzero}—by treating the correct choice as ground truth, thereby facilitating simple and robust reward modeling.

A key challenge in training judgment models lies in obtaining high-quality candidate responses with a clearly known best answer. This is particularly difficult for multimodal large language models (MLLMs), as even state-of-the-art models exhibit limited judgment capabilities~\citep{li2024vlrewardbenchchallengingbenchmarkvisionlanguage}. As a result, generating well-scored, challenging response candidates remains non-trivial. To address this, we take an alternative approach termed reverse response candidates synthesis, where we intentionally generating lower-quality negative responses, which is more tractable. Starting with a set of visual supervised fine-tuning (SFT) datasets as seeds, we treat the original annotations as positive references and prompt the MLLM to generate plausible yet imperfect alternatives. This strategy enables the creation of diverse and challenging candidate sets without relying on external supervision or additional human labeling. Importantly, our method does not assume the original annotations are flawless; since the objective is to learn relative preferences by degrading response quality, absolute correctness is not required for effective training.

Although with the annotated candidate responses, we can already tune the MLLM judge using RL from scratch, we observe limited emergence of complex reasoning behaviors such as re-evaluation or deliberate reflection moments. We hypothesize that this stems from MLLMs' thorough supervised fine-tuning (SFT) stage, which enhances output structure but reduces diversity, consequently restricting exploration during RL. To address this, we carefully design a synthetic data generation pipeline to extract the long-reasoning capability from text-based reasoning LLMs (e.g., DeepSeek-R1~\citep{deepseekai2025deepseekr1incentivizingreasoningcapability}), which enables the judge model to regain its reasoning strength via fine-tuning.

Our training pipeline follows a two-stage procedure. First, we leverage the extracted reasoning traces and apply SFT to warm up the MLLM with reasoning ability. In the second stage, we apply RL to refine the model’s ability to leverage its reasoning capability to improve its judgment for responses across diverse visual tasks.

In summary, our contributions are as follows: 
\begin{itemize} 
\item We propose a novel formulation for training MLLM-as-judges, which is cast as a multiple-choice problem with explicit reasoning process. 
\item We introduce a novel scalable annotation strategy to construct response candidates with known best answer without relying on external scoring models or human annotations. 
\item We carefully design a pipeline for constructing SFT dataset that extracts the strong reasoning capability from text-based reasoning LLMs, which enables the judge model to unlock its reasoning strength.
\item With our proposed data annotation strategy, we curate \Data containing around 3K long thought annotations for SFT and 5K multiple-choice QA for RL.
\item We demonstrate through extensive experiments that our approach significantly improves the MLLM judge's performances for various multimodal tasks. Specifically, our \ModelNameAbbre-7B surpasses GPT-4o by 9.9\% on VL-RewardBench~\citep{li2024vlrewardbenchchallengingbenchmarkvisionlanguage} and improves performance on MM-Vet during inference-time scaling by up to 7.7\%.
\end{itemize}

\section{Related Work}

\paragraph{Multi-Modal Large Language Model.}
Recent advancements in large language models (LLMs) have significantly improved language comprehension and generation, achieving near-human proficiency across various tasks~\citep{brown2020language, scao2022bloom, chowdhery2022palm, smith2022using, hoffmann2022training, ouyang2022training, touvron2023llama, bai2022training, zhang2025vlm2benchcloserlookvlms}. This success has spurred interest in vision-language interaction, leading to multi-modal large language models (MLLMs)~\citep{liu2023llava, li2023blip2, dai2023instructblip, zhu2023minigpt4, dai2023instructblip, openai2023gpt4, bai2023qwenvl, su2023pandagpt, gao2023llamaadapter, pi2023detgpt, pi2023perceptiongpt, pi2024strengthening, pi2024mllmprotector, gao2023gllava}, which excel in dialogue based on visual inputs. 

\paragraph{LLM as Judge and Reward Modeling}
Previous work has explored reward models and LLM-as-Judges to align large models with human preferences. Early approaches used the Bradley-Terry formulation to assign scalar scores to responses~\citep{ouyang2022traininglanguagemodelsfollow}. PairRM~\cite{jiang2023llmblenderensemblinglargelanguage} improved this by comparing response pairs to capture relative quality. More recently, LLM-as-Judges have emerged as a strong alternative~\citep{zheng2023judgingllmasajudgemtbenchchatbot, yuan2025selfrewardinglanguagemodels, guo2024directlanguagemodelalignment, liu2025inferencetimescalinggeneralistreward}, offering both rankings and decision rationales. In vision-language tasks, reward models have been adapted to incorporate human preferences, improving text-to-image generation~\cite{lee2023image}, open-ended image reasoning~\cite{yuan2023scaling}, and general visual feedback modeling~\cite{dong2023unified}. Nonetheless, building effective judges for MLLMs remains an open challenge.

\paragraph{Reasoning Models} 
Following the introduction of OpenAI’s O1 model series~\citep{openai2024openaio1card}, a growing body of research has focused on scaling the inference-time compute of large language models (LLMs) to enhance performance~\citep{snell2024scalingllmtesttimecompute, wu2025inferencescalinglawsempirical, kumar2024traininglanguagemodelsselfcorrect}. DeepSeek-R1 has recently demonstrated the effectiveness of rule-based reinforcement learning (RL) in enabling LLMs to perform complex reasoning~\citep{lu2024deepseek}. Concurrently, several studies have sought to extend such reasoning capabilities to multimodal LLMs (MLLMs)~\citep{yang2025r1onevisionadvancinggeneralizedmultimodal, meng2025mmeurekaexploringfrontiersmultimodal, shen2025vlmr1stablegeneralizabler1style, xu2025llavacotletvisionlanguage}, verifying the feasibility of reasoning over visual inputs. However, the potential of leveraging multimodal reasoning models as evaluative judges is still under-explored, and whether such enhancement improves the judgment performance is yet to be examined.
\section{Problem Formulation}
We formulate the judgement process as a multiple-choice selection task. Specifically, the MLLM judge first conducts reasoning based on the candidate responses by conducting detailed analysis. Then, it makes a conclusion by selecting the best candidate response from the given ones. Each candidate response is associated with a label (e.g. A,B,C).

Let $\II$ be the multimodal inputs (text, image, etc.) and $\RR = \{r_1, \ldots, r_K\}$ be a set of $K$ candidate responses, where each $r_k$ is associated with a label $\ell_k \in \LL = \{A, B, \ldots\}$.
The MLLM judge $J$ operates in two parts:

\textbf{Structured Reasoning}
The judger first generates a structured analysis:
\begin{equation}
    \text{Analysis} = J_{\text{think}}(\phi (\II, \RR)) = \texttt{<think>}\TT\texttt{</think>}
\end{equation}
where $\TT$ represents the reasoning trace evaluating each response's quality. $\phi$ is a template function that composes the inputs and responses into a multiple choice question. Detailed template is demonstrated in Table~\ref{tab:mc_template}.

\textbf{Response Selection}
The judger outputs its final selection:
\begin{equation}
    \text{Selection} = J_{\text{answer}}(\phi (\II, \RR), \TT) = \texttt{<boxed>\{}\ell^*\texttt{\}}
\end{equation}
where $\ell^* \in \LL$ is the predicted best response label.

The reasoning-enhanced multiple-choice formulation offers several key benefits: (1) it enables rule-based reinforcement learning by constraining the output space to discrete choices, simplifying both reward assignment and policy optimization; (2) evaluation becomes straightforward, as correctness reduces to exact label matching, eliminating the need for subjective scoring via an auxiliary reward model; and (3) it naturally encourages structured reasoning, as the model must perform comparative analysis across candidates before making a selection. Additionally, the generated reasoning traces can lead to more informed and accurate judgments.
\begin{table*}[t!]\centering
\begin{minipage}{1.0\textwidth}\vspace{0mm}    \centering
\begin{sectionbox}[]{Prompt for Negative Candidates Generation}
    \centering
      \footnotesize
    \begin{tabular}{p{0.97\textwidth} c}
I am providing you with an image, a question, and a correct answer. You are an expert in injecting errors into the correct answer. Consider the following types of errors: \\
1) hallucination, which reflects inaccuracy in the response that is caused by misinterpreting the image, such as object existence, spatial relationship, object attributes, or misreading texts from the image (OCR) \\
2) incompleteness, where the response is too simple and does not cover all the important information \\
3) incorrect reasoning, where the response contains reasoning that is not truthful (wrong calculation, wrong geometric reasoning, wrong common knowledge) \\
4) incorrect knowledge, where the response contains factual knowledge that is incorrect, i.e., the function some tool, the purpose/effect of some well known action \\
\\
You are responsible for the following:\\
1) Identify what error is most likely to occur given the image and the question.\\
2) Modify the correct answer, such that the modified answer is reasonable but contains the error.\\
3) Output both the modified answer and error type.\\
\\
You must output the answer in the following format:\\
$<$think$>$thinking process of what error is appropriate to add$<$/think$>$\\
$<$error type$>$error[e.g., hallucination]$<$/error type$>$\\
$<$error detail$>$details of the added error$<$/error detail$>$\\
$<$modified answer$>$modifed answer$<$/modified answer$>$\\
\\
Here are some examples, where the images are represented using captions: \\
$<$examples$>$\\
Here are the image, question and correct answer:\\
image: $<$image$>$\\
question: \{question\}\\
answer: \{answer\}\\
    \end{tabular}
\end{sectionbox}
\vspace{-5mm}
\caption{Prompt for generating negative responses based on seed datasets.}
    \label{tab:neg_prompt}
\end{minipage}
\end{table*}

\section{Response Candidate Construction}\label{sec: neg_gen}
A critical challenge in training \ModelNameAbbre is obtaining high-quality candidate responses with reliable best-answer annotations. While one could employ external state-of-the-art MLLMs for scoring and annotation~\citep{li2023silkie}, this approach faces two limitations: limited accessibility to such models, and potential unreliability in their evaluations.

\paragraph{Reverse Response Candidates Synthesis}
We address the challenge of generating labeled response sets through a self-annotation strategy that repurposes existing supervised fine-tuning (SFT) data. Our approach operates in reverse: the original annotated responses from the SFT dataset are treated as the reference ``best” candidates, while negative candidates are synthesized by prompting the MLLM to introduce carefully designed errors. These alternatives remain contextually relevant but contain subtle flaws, making them plausible yet suboptimal. An example prompt used to generate such negative responses is shown in Table~\ref{tab:neg_prompt}.
A natural concern is the quality of the original SFT annotations. Crucially, our method does not assume the reference responses are perfect. The only requirement is that the synthesized negatives be of relatively lower quality than the reference. This aligns seamlessly with our multiple-choice formulation, where relative ranking among candidates is sufficient and absolute correctness is unnecessary. As a result, the framework is robust to imperfections in the original data while still supporting effective training through contrastive supervision.

\paragraph{Multiple-Choice Formatting}
After obtaining the candidate responses, we use a template to format the input image, input text query and candidate responses in to a multiple choice problem. The template is shown in Table~\ref{tab:mc_template}. Specifically, in the prompt, we specify the criteria for evaluating the responses. We consider three aspects for ranking: 1) harmfulness, which reflects whether the response aligns with human value, and does not contain malicious contents; 2) accuracy: whether the responses contains any hallucination, and present correct attributes; 3) detailedness: the response with more details should be preferred. The importance weighting when comparing the responses is: harmfulness $>$ accuracy $>$ detailedness. This criteria is specified and designed based on the conventional way of comparing responses. Customization can be made to adapt to new scenarios.

\textbf{Variable Number of Candidates}
When constructing multiple-choice questions, we vary the number of candidate responses from 2 to 4. Allowing a flexible number of candidates enables the judge to evaluate more than two responses simultaneously during inference, improving efficiency. Moreover, training the judge with varying numbers of options (e.g., mixing 2-choice and 4-choice questions within the same batch) helps prevent overfitting to a fixed candidate count and enhances the model's generalization and robustness across different application settings.

\textbf{Candidate Order Shuffling} We randomize both the order of candidate responses and the label assigned to the best response (e.g., rotating the "A/B/C/D" designations) to prevent the judge from learning positional or label-based shortcuts. Empirical results demonstrate that this shuffling is essential for ensuring robust and unbiased performance.

\section{Improving MLLM Judge with Reinforcement Learning}
Inspired by recent works on improving reasoning capabilities of LLMs via reinforcement learning, we adopt a similar approach to strengthen the ability of \ModelNameAbbre. 
The composite reward is a weighted combination of format reward and accuracy reward. The former encourages the judge to output the contents in our specified format, which facilitates extraction of reasoning process and final selection. The latter encourages the judge to select the right response. Formally, the two reward components can be defined as follows:
\vspace{-3.0mm}
\begin{equation}
    R_{\text{format}} = 
    \begin{cases}
        1.0 & \text{if all the tags are correctly formatted}\\
        0.0 & \text{if formatting errors exist}
    \end{cases}
\end{equation}
\begin{equation}
    R_{\text{accuracy}} = 
    \begin{cases} 
        1.0 & \text{if } \ell^* = \ell^{\text{gt}}\\
        0.0 & \text{otherwise}
    \end{cases}
\end{equation}
The total reward is computed as:
\begin{equation}
    R_{\text{total}} = (1-\alpha) \cdot R_{\text{accuracy}} + \alpha \cdot R_{\text{format}}
\end{equation}
$\alpha$ is the weighting term for format reward, which we set 0.1 in our experiments. The overall training objective becomes:
\vspace{-2mm}
\begin{equation}
    \theta^* = \arg\max_\theta \mathbb{E}_{(\II,\RR)}\left[R_{\text{total}}(J_\theta(\II,\RR))\right]
\end{equation}\label{eqn:total_reward}
We employ Group Relative Policy Optimization (GRPO)~\citep{shao2024deepseekmathpushinglimitsmathematical} as our optimization algorithm, favoring it over PPO due to its enhanced stability and reduced computational demands. GRPO eliminates the need for a separate value function (critic) by utilizing group-based advantage estimation, where multiple responses are generated for each prompt, and their rewards are normalized to compute advantages. This approach not only simplifies the training process but also significantly reduces memory usage and computational overhead. 


\section{Waking \ModelNameAbbre's Long-thought Reasoning}

While direct reinforcement learning (RL) application to MLLMs can already improve performance, we observe limited emergence of complex reasoning behaviors such as re-evaluation. We hypothesize that this stems from MLLMs' thorough supervised fine-tuning (SFT) stage, which enhances output structure but reduces diversity, consequently restricting exploration during RL. To address this, we first revive the MLLM's exploration capability via supervised fine-tuning using data containing extended reasoning traces generated by text-based reasoning models.

\subsection{Extract Reasoning Ability from Text-based Reasoner}
Our SFT dataset is synthesized based on the previously generated MC-questions (described in Section~\ref{sec: neg_gen}) by enriching the response with detailed reasoning and reflection. However, the current SOTA MLLMs are not able to produce reliable complex reasoning traces. Therefore, we turn to text-based reasoning models, and conduct the following phases to distill their reasoning capability:

\paragraph{Modality Bridging via Image Description} For each image, we generate a detailed textual description using an MLLM, which serves as a bridge to effectively translate visual information into a language format. This allows text-only models such as DeepSeek-R1 to reason about and evaluate visual content without direct access to the image itself. 

\begin{table*}[t!]
\caption{Results on VL-RewardBench~\citep{li2024vlrewardbenchchallengingbenchmarkvisionlanguage}. We report accuracies for all subcategories of the benchmark. Overall Accuracy is the point-wise average accuracy of all questions and answer pairs. Macro Accuracy is the average taken over three subcategories. We observe that our \ModelNameAbbre variants achieve strong performances across all subcategories, and even rival with proprietary models.}
\vspace{-2mm}
\label{tab:vl_rewardbench_results}
\centering
\resizebox{\textwidth}{!}{
\begin{tabular}{l!{\vrule width 0.5pt}ccc!{\vrule width 0.5pt}cc}
\toprule
\textbf{Models} & \textbf{General} & \textbf{Hallucination} & \textbf{Reasoning} & \textbf{Overall Accuracy} & \textbf{Macro Accuracy} \\
\midrule
\multicolumn{6}{c}{\textit{Open-Source Models}} \\
\midrule
LLaVA-OneVision-7B-ov & 32.2 & 20.1 & 57.1 & 29.6 & 36.5 \\
InternVL2-8B         & 35.6 & 41.1 & 59.0 & 44.5 & 45.2 \\
Phi-3.5 Vision       & 28.0 & 22.4 & 56.6 & 28.2 & 35.7 \\
Qwen2-VL-7B         & 31.6 & 19.1 & 51.1 & 29.3 & 33.9 \\
Qwen2-VL-72B        & 38.1 & 32.8 & 58.0 & 39.5 & 43.0 \\
Llama-3.2-11B       & 33.3 & 38.4 & 60.8 & 42.9 & 43.7 \\
Llama-3.2-90B       & 42.6 & 57.3 & 61.7 & 56.2 & 53.9 \\
Molmo-7B            & 31.1 & 19.4 & 55.7 & 31.7 & 35.2 \\
Molmo-72B           & 33.9 & 42.3 & 54.9 & 44.1 & 43.7 \\
Pixtral-12B         & 28.5 & 25.9 & 59.4 & 35.8 & 38.0 \\
NVLM-D-72B          & 38.9 & 31.6 & 62.0 & 40.1 & 44.1 \\
\midrule
\multicolumn{6}{c}{\textit{Proprietary Models}} \\
\midrule
Claude-3.5 Sonnet (2024-06-22) & 43.4 & 50.5 & 62.3 & 55.3 & 53.6 \\
GPT-4o-mini (2024-07-18)      & 41.7 & 34.5 & 58.2 & 41.5 & 44.8 \\
GPT-4o (2024-08-06)           & 49.1 & 67.6 & \textbf{\underline{70.5}} & 65.8 & 62.4 \\
\midrule
\multicolumn{6}{c}{\textit{Judge Models}} \\
\midrule
LLaVA-Critic-7B           & 47.7 & 45.5 & 58.7 & 49.6 & 50.6 \\
CAREVL-Qwen2VL-7B           & - &  - & 56.3 & 67.8 & - \\
CAREVL-OneVision-7B           & - &  - & 61.4 & 68.7 & - \\
CAREVL-LLaMA-11B           & - &  - & 60.8 & 70.7 & - \\
XC-2.5-Reward           & - & - & \textbf{62.9} & 65.8 & - \\
\midrule
\cc \textbf{\ModelNameAbbre-3B-SFT} & \cc 59.6 & \cc 78.9 & \cc 52.8 & \cc 69.4 & \cc 63.8 \\
\cc \textbf{\ModelNameAbbre-3B-Zero-RL} & \cc \textbf{66.1} & \cc 78.8 & \cc 52.5 & \cc 70.2 & \cc 65.8 \\
\cc \textbf{\ModelNameAbbre-3B-SFT-RL} & \cc \underline{65.0} & \cc \underline{81.6} & \cc 54.1 & \cc \underline{72.2} & \cc \textbf{66.9} \\
\midrule
\cc \textbf{\ModelNameAbbre-7B-SFT} & \cc 55.2 & \cc 79.8 & \cc 56.6 & \cc 70.3 & \cc 68.9 \\
\cc \textbf{\ModelNameAbbre-7B-Zero-RL} & \cc 60.2 & \cc \textbf{82.5} & \cc \textit{56.6} & \cc \textbf{72.7} & \cc \underline{66.4} \\
\cc \textbf{\ModelNameAbbre-7B-SFT-RL} & \cc \textbf{\underline{68.7}} & \cc \textbf{\underline{83.2}} & \cc \underline{61.4} & \cc \textbf{\underline{75.5}} & \cc \textbf{\underline{71.1}} \\
\bottomrule
\end{tabular}
}
\end{table*}

\paragraph{Reasoning Trace Generation with Hinted Prompting} We employ a text-based reasoning model—e.g., DeepSeek-R1-distilled Qwen—to generate reasoning traces that justify the selection of the best response. Since image descriptions alone may occasionally lack sufficient detail for reliable decision-making, we include a subtle hint indicating the correct answer to guide the generation of reasoning traces. The LLM is prompted to reason as if making the evaluation independently.

\paragraph{Hint Reference Removal and Style Alignment} Post-generation analysis revealed that hint references (e.g., ``as the hint suggests...") may still exist despite explicit instructions to avoid them. In addition, since image descriptions are used to represent images when creating reasoning traces, the reasoning model usually refers to the images using phrases like ``as stated in the image description". We therefore implement an LLM-based cleaning process to: (1) remove all hint references, and (2) convert ``image description" terminology to direct ``image" references, ensuring no information is leaked during training, and the output format is consistent with the target MLLM judge's expected style.

\subsection{Truncated Reward Assignment}
After the warm-up stage using our curated long-form reasoning data, we observe an intriguing phenomenon: although the average response length increases significantly following supervised fine-tuning (SFT), continued reinforcement learning (RL) of the judge MLLM yields only marginal performance improvements. Analyzing the validation reward curves for both \( R_{\text{accuracy}} \) and \( R_{\text{format}} \), we find that while the former improves steadily, the latter plateaus or even deteriorates after several RL steps (as illustrated in Figure~\ref{fig:trunc}). To better understand this behavior, we examine responses that receive a score of zero for \( R_{\text{format}} \) and identify two predominant failure modes: (1) repeated outputs and (2) excessively long or never-ending reasoning. We hypothesize that these issues stem from the MLLM's limited exposure to long-form responses during its pretraining phase. To mitigate this, we propose incorporating a length constraint into the reward formulation in Equation~\ref{eqn:total_reward}, which becomes the following: 
\begin{equation}
R_{\text{total}} =
\begin{cases}
0, \quad \text{if length(response)} > L \\
(1-\alpha) \cdot R_{\text{accuracy}} + \alpha \cdot R_{\text{format}}, \quad \text{otherwise}
\end{cases}
\end{equation}
where $L$ is the max length of the response. We empirically find that setting the maximum length to 1024 achieves the best performance. After applying this revised reward assignment, we observe that the validation reward successfully increases with additional RL steps.

\section{Experiments}
In this section, we demonstrate the effectiveness of \ModelNameAbbre trained with our \Data and training pipeline.

\paragraph{Evaluation Benchmarks} For evaluation, we mainly adopt VL-RewardBench~\citep{li2024vlrewardbenchchallengingbenchmarkvisionlanguage}, which offers a broad evaluation framework aimed at assessing diverse multimodal queries. 
We re-formulate the samples into MC questions with 2 candidate responses, and exact match is used to identify whether the judgement is correct. We report the accuracies for this benchmark. For inference time scaling experiment, we select MM-Vet~\citep{yu2023mmvet} and LLaVA-Bench~\citep{liu2023llava} as our task benchmark, which covers diverse real world scenarios and the responses are freeform. The evaluation is performed by GPT4V with reference ground truth responses.

\paragraph{Baselines}
We compare \ModelNameAbbre with a wide array models. Specifically, we compare with open-source MLLMs including LLaVA-OneVision~\citep{li2024llavaonevisioneasyvisualtask}, InterVL~\citep{chen2024internvl}, Phi-3.5VL~\citep{abdin2024phi3technicalreporthighly}, Qwen2VL~\citep{bai2023qwenvl}, LLama3.2~\citep{touvron2023llama}, Molmo~\citep{deitke2024molmopixmoopenweights}, Pixtral~\citep{agrawal2024pixtral12b} and NVLM~\citep{dai2024nvlmopenfrontierclassmultimodal}. We also compare with proprietary models including Claude-3.5~\citep{TheC3} and GPT4o~\citep{openai2024gpt4ocard}. The results are derived from VL-RewardBench~\citep{li2024vlrewardbenchchallengingbenchmarkvisionlanguage}. We also compare with specially trained MLLM judges, including LLaVA-Critic~\citep{xiong2024llava}, CAREVL~\citep{dai2025carevl} and XC-2.5-Reward~\citep{zang2025internlm}. Note that since the results reported by \cite{dai2025carevl} \cite{zang2025internlm} are calculated by an outdated version of the script, where parts of the ``General" and ``Hallucination" are mixed up, we only report their ``Reasoning" and ``Overall Accuracy" results.

\paragraph{Training Data Sources} We curate \Data using approximately 100,000 multimodal SFT samples collected from diverse sources, including Allava~\cite{zhang2024allava}, AI2D~\cite{ai2d}, Chart2Text~\cite{kantharaj2022chart}, ChartQA~\cite{masry2022chartqa}, CLEVR~\cite{johnson2017clevr}, CLEVR-Math~\cite{abraham2022clevr}, diagram-i2t~\cite{diagrami2t}, DVQA~\cite{kafle2018dvqa}, FigureQA~\cite{kahou2018figureqa}, Geo170k~\cite{gao2023gllava}, Geo3K~\cite{geo3k}, Geos~\cite{geos2024}, ScienceQA~\cite{lu2022learn}, and TextOCR~\cite{singh2021textocr}. For each seed sample, we generate four negative candidate responses and apply our proposed negative sampling strategy. After post-processing and cleaning, the dataset is split into two subsets: 31,703 samples for SFT and 52,080 for RL. Long-form reasoning traces are then created for the SFT subset.

\paragraph{Implementation Details}
We adopt Qwen2.5VL~\cite{bai2023qwenvl} models as our base MLLM. For warm up SFT, we use adopt learning rate $1\times10^{-5}$ to conduct finetuning. We use AdamW optimizer with cosine learning rate schedule. For RL, we use 512 as roll out batch size, training batch size is set to 128. Temperature is set to 1.0 during roll out, and the number of generated responses for each question is set to 5.
\subsection{Evaluation on VL-RewardBench}
In Table~\ref{tab:vl_rewardbench_results}, we demonstrate the effectiveness of our MLLM judge using VL-RewardBench~\citep{li2024vlrewardbenchchallengingbenchmarkvisionlanguage}. The results of the baseline MLLMs are derived from the original papers. We observe that our MLLM judge initialized from small MLLMs can rival with larger MLLMs, and even proprietary MLLMs on this benchmarks. In addition, we observe that even though the judge model directly tuned with RL already achieves considerable performance gain, conducting the warm up stage by distilling from reasoning LLMs still achieve considerable performance boost.

\begin{table*}[t!]
\caption{Results on MM-Vet~\citep{yu2023mmvet} for MLLM inference-time scaling. We infer the task model 4 times for each question, and use \ModelNameAbbre-7B to select the most promising response. We observe substantial performance gains across all sub tasks. Interestingly, weaker task MLLMs typically achieve more gain after applying \ModelNameAbbre.}
\label{tab:mmvet}
\vspace{-2mm}
\centering
\resizebox{\textwidth}{!}{
\begin{tabular}{l!{\vrule width 0.5pt}cccccc!{\vrule width 0.5pt}c}
\toprule
\textbf{Model} & \textbf{Recognition} & \textbf{OCR} & \textbf{Knowledge} & \textbf{Generation} & \textbf{Spatial} & \textbf{Math}& \textbf{Total} \\
\midrule
\textbf{OneVision-0.5B} & 47.9 & 55.5 & 34.6 & 38.9 & 49.3 & 41.9 & 50.6 \\
\cc \textbf{OneVision-0.5B+\ModelNameAbbre} & \cc 51.3\textsuperscript{\(\uparrow 3.4\)} & \cc 62.0\textsuperscript{\(\uparrow 6.5\)} & \cc 37.3\textsuperscript{\(\uparrow 2.7\)} & \cc 42.3\textsuperscript{\(\uparrow 3.4\)} & \cc 56.1\textsuperscript{\(\uparrow 6.8\)} & \cc 55.4\textsuperscript{\(\uparrow 13.5\)} & \cc 54.5\textsuperscript{\(\uparrow 3.9\)} \\
\midrule
\textbf{OneVision-7B} & 47.9 & 55.9 & 34.8 & 39.4 & 49.7 & 42.3 & 50.8 \\
\cc \textbf{OneVision-7B+\ModelNameAbbre} & \cc 51.3\textsuperscript{\(\uparrow 3.4\)} & \cc 62.2\textsuperscript{\(\uparrow 6.3\)} & \cc 37.3\textsuperscript{\(\uparrow 2.5\)} & \cc 42.3\textsuperscript{\(\uparrow 2.9\)} & \cc 56.1\textsuperscript{\(\uparrow 6.4\)} & \cc 55.4\textsuperscript{\(\uparrow 13.1\)} & \cc 54.6\textsuperscript{\(\uparrow 3.8\)} \\
\midrule
\textbf{InterVL-2B} & 48.5 & 56.0 & 32.7 & 35.5 & 50.7 & 50.0 & 52.3 \\
\cc \textbf{InterVL-2B+\ModelNameAbbre} & \cc 55.9\textsuperscript{\(\uparrow 7.4\)} & \cc 65.9\textsuperscript{\(\uparrow 9.9\)} & \cc 41.4\textsuperscript{\(\uparrow 8.7\)} & \cc 44.6\textsuperscript{\(\uparrow 9.1\)} & \cc 61.6\textsuperscript{\(\uparrow 10.9\)} & \cc 60.8\textsuperscript{\(\uparrow 10.8\)} & \cc 60\textsuperscript{\(\uparrow 7.7\)} \\
\midrule
\textbf{InterVL-8B} & 49.5 & 60.8 & 33.3 & 36.4 & 60.5 & 53.8 & 53.9 \\
\cc \textbf{InterVL-8B+\ModelNameAbbre} & \cc 53.7\textsuperscript{\(\uparrow 4.2\)} & \cc 66.8\textsuperscript{\(\uparrow 6.0\)} & \cc 43.5\textsuperscript{\(\uparrow 10.2\)} & \cc 47.7\textsuperscript{\(\uparrow 11.3\)} & \cc 62.7\textsuperscript{\(\uparrow 2.2\)} & \cc 53.8\textsuperscript{\(\uparrow 0\)} & \cc 58.5\textsuperscript{\(\uparrow 4.6\)} \\
\midrule
\textbf{Qwen2.5-3B} & 48.2 & 57.0 & 33.6 & 35.0 & 51.9 & 50.0 & 52.6 \\
\cc \textbf{Qwen2.5-3B+\ModelNameAbbre} & \cc 55.4\textsuperscript{\(\uparrow 7.2\)} & \cc 65.7\textsuperscript{\(\uparrow 8.7\)} & \cc 40.7\textsuperscript{\(\uparrow 7.1\)} & \cc 43.7\textsuperscript{\(\uparrow 8.7\)} & \cc 61.5\textsuperscript{\(\uparrow 9.6\)} & \cc 60.8\textsuperscript{\(\uparrow 10.8\)} & \cc 59.6\textsuperscript{\(\uparrow 7.0\)} \\
\midrule
\textbf{Qwen2.5-7B} & 62.1 & 68.5 & 54.3 & 57.2 & 62.4 & 58.1 & 64.0 \\
\cc \textbf{Qwen2.5-7B+\ModelNameAbbre} & \cc 65.0\textsuperscript{\(\uparrow 2.9\)} & \cc 72.6\textsuperscript{\(\uparrow 4.1\)} & \cc 57.4\textsuperscript{\(\uparrow 3.1\)} & \cc 60.4\textsuperscript{\(\uparrow 3.2\)} & \cc 67.9\textsuperscript{\(\uparrow 5.5\)} & \cc 73.1\textsuperscript{\(\uparrow 15.0\)} & \cc 67.8\textsuperscript{\(\uparrow 3.8\)} \\
\bottomrule
\end{tabular}
}
\end{table*}

\begin{figure}[h]
\centering
\includegraphics[width=0.47\textwidth]{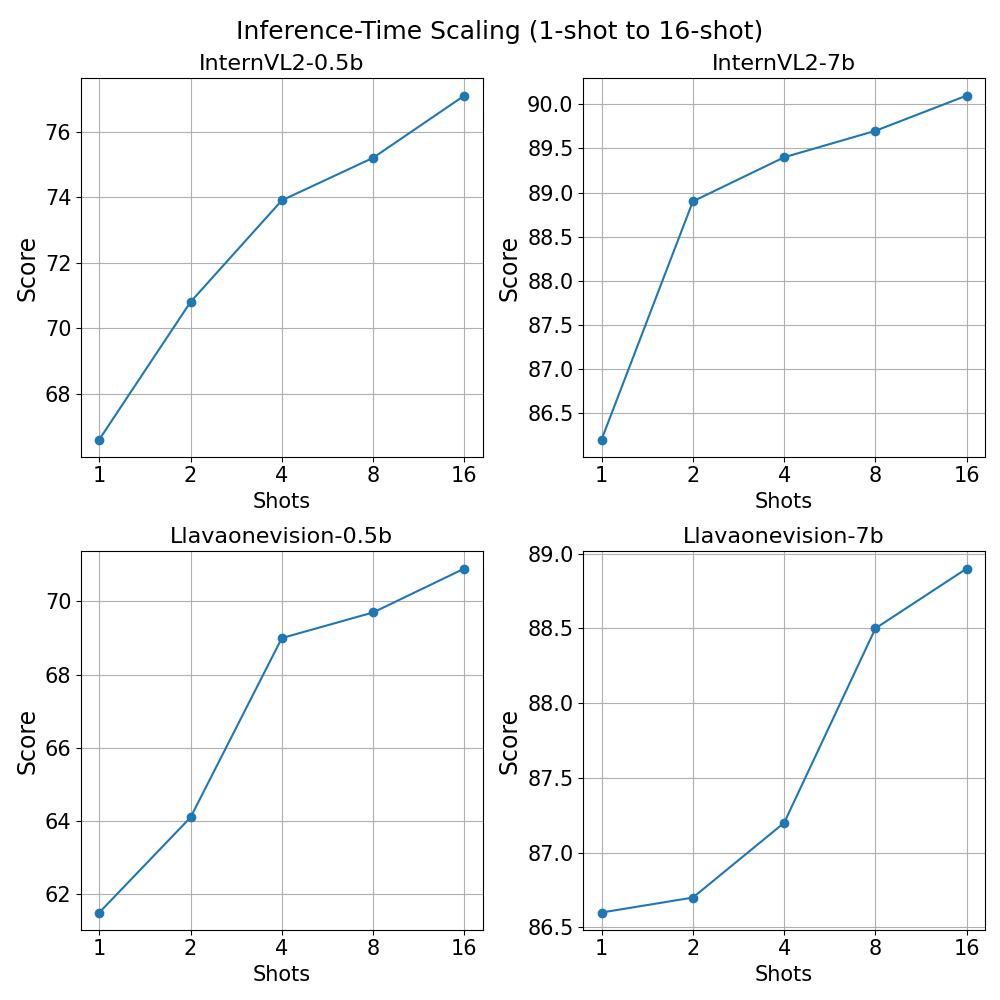} 
\vspace{-0.2cm}
\caption{Inference time scaling on LLaVA-Bench.
}\label{fig:inference_scaling_llavabenc}
\end{figure}
\subsection{Inference-time Scaling for MLLMs}
We demonstrate the effectiveness of \ModelNameAbbre in enhancing MLLM performance via inference-time scaling, evaluated on two diverse free-form VQA benchmarks: MM-Vet~\citep{yu2023mmvet} and LLaVA-Bench~\citep{liu2023llava}. As shown in Table~\ref{tab:mmvet}, we generate four diverse responses per question on MM-Vet using a temperature of 0.9, convert them into a multiple-choice format, and use \ModelNameAbbre-7B to select the best answer. Results consistently show improved performance across model families and question types, highlighting \ModelNameAbbre's effectiveness in candidate selection. Figure~\ref{fig:inference_scaling_llavabenc} shows performance on LLaVA-Bench improves as more responses are sampled. Responses are recursively paired, reformulated into binary multiple-choice questions, and evaluated by \ModelNameAbbre-7B until a final answer is chosen. This confirms that inference-time scaling with \ModelNameAbbre leads to steady performance gains.



\begin{table*}[t!]
\caption{Majority voting improves \ModelNameAbbre. We infer the judge model multiple times, which produces diverse reasoning traces and judgments, then we perform majority voting to derive the final selection.}
\label{tab:majority_voting}
\centering
\vspace{-2mm}
\resizebox{\textwidth}{!}{
\begin{tabular}{l!{\vrule width 0.5pt}ccc!{\vrule width 0.5pt}cc}
\toprule
\textbf{Models} & \textbf{General} & \textbf{Hallucination} & \textbf{Reasoning} & \textbf{Overall Accuracy} & \textbf{Macro Average Accuracy} \\
\midrule
\textbf{\ModelNameAbbre-3B-SFT-RL} & 65.0 & 81.6 & 54.1 & 72.2 & 66.9 \\
\cc @5-shot & \cc 71.0\textsuperscript{\(\uparrow 6.0\)} & \cc 84.3\textsuperscript{\(\uparrow 2.7\)} & \cc 57.8\textsuperscript{\(\uparrow 3.7\)} & \cc 75.6\textsuperscript{\(\uparrow 3.4\)} & \cc 71.1\textsuperscript{\(\uparrow 4.2\)} \\
\cc @10-shot & \cc 72.7\textsuperscript{\(\uparrow 7.7\)} & \cc 85.2\textsuperscript{\(\uparrow 3.6\)} & \cc 57.6\textsuperscript{\(\uparrow 3.5\)} & \cc 76.3\textsuperscript{\(\uparrow 4.1\)} & \cc 71.8\textsuperscript{\(\uparrow 4.9\)} \\
\midrule
\textbf{\ModelNameAbbre-7B-SFT-RL} & 68.7 & 83.2 & 61.4 & 75.5 & 71.1 \\
\cc @5-shot & \cc 69.2\textsuperscript{\(\uparrow 0.5\)} & \cc 84.8\textsuperscript{\(\uparrow 1.6\)} & \cc 63.4\textsuperscript{\(\uparrow 2.0\)} & \cc 77.1\textsuperscript{\(\uparrow 1.6\)} & \cc 72.5\textsuperscript{\(\uparrow 1.4\)} \\
\cc @10-shot & \cc 70.3\textsuperscript{\(\uparrow 1.6\)} & \cc 87.5\textsuperscript{\(\uparrow 4.3\)} & \cc 65.2\textsuperscript{\(\uparrow 3.8\)} & \cc 79.5\textsuperscript{\(\uparrow 4.0\)} & \cc 74.3\textsuperscript{\(\uparrow 3.2\)} \\
\bottomrule
\end{tabular}
}
\end{table*}
\subsection{Improving \ModelNameAbbre via Majority Voting}
Thanks to the generative nature of \ModelNameAbbre, its performance can be further enhanced by increasing the number of inference passes. This improvement occurs not only because repeated inference allows MLLMs to generate more diverse candidate responses from which the best can be selected, but also because \ModelNameAbbre itself becomes more reliable when allowed to sample multiple times and apply majority voting over its own predictions. This dual benefit highlights the scalability of \ModelNameAbbre under increased compute budgets. As demonstrated in Table~\ref{tab:majority_voting}, we observe a significant performance gain on VL-RewardBench when increasing the number of inference samples. For this experiment, we set the temperature of the judge model to 0.9 to encourage diverse generations, which in turn enriches the voting pool and improves overall decision quality.

\begin{figure}[h]
\includegraphics[width=0.49\textwidth]{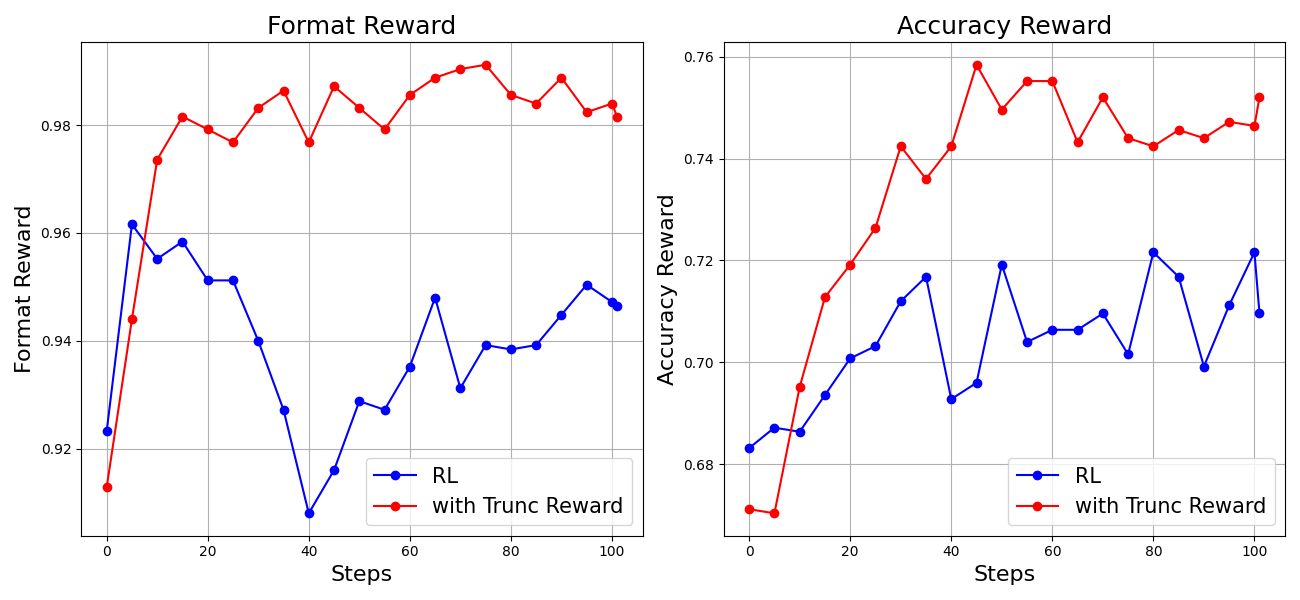} 
\vspace{-6mm}
\caption{The validation reward curves of original RL and RL with truncated reward assignment.
}\label{fig:trunc}
\end{figure}
\subsection{Impact of Truncated Reward Assignment}
We demonstrate the curves of accuracy reward and validation reward calculated on the validation set in Figure~\ref{fig:trunc}. We observe that for the original RL reward assignment, the format reward deteriorates after a few steps of optimization. This is because during RL, the reward assignment often favors longer responses, which leads the judge model to generating longer and longer responses. However, during the pre-training stage of the original MLLM, the long context pretraining data only takes up a small proportion, which makes the MLLMs less adept at handling excessively long responses. After applying the truncated reward assignment, both the format reward and the accuracy reward begin to steadily climb.

\section{Conclusion}
We propose \ModelNameAbbre, a reasoning-centric framework that empowers general-purpose MLLM judges to deliver more accurate and interpretable evaluations. By reformulating judgment as a multiple-choice problem grounded in deliberate reasoning, and by introducing scalable strategies for data synthesis—including response candidate generation and reasoning distillation—our method equips MLLMs with enhanced evaluative capabilities. Extensive experiments validate the effectiveness of \ModelNameAbbre, with significant improvements over strong baselines such as GPT-4o on VL-RewardBench and consistent gains on MM-Vet through inference-time scaling. We hope this work provides insights for future research into more advanced and powerful MLLM judges, encouraging the community to explore richer reasoning strategies, more robust supervision signals, and principled evaluation protocols for aligning model judgments with human preferences.

\section{Limitation}
Despite the impressive performance of \ModelNameAbbre, it requires generating a lengthy chain of reasoning before producing a final judgment. While this reasoning trace offers valuable benefits—such as improved interpretability and reduced vulnerability to reward hacking—it also increases evaluation time. This, in turn, raises the computational cost for both RLHF training and inference-time scaling. As a promising direction for future work, it is worth exploring strategies to shorten the output length without compromising judgment quality.

\bibliography{main}

\begin{thebibliography}{79}
\expandafter\ifx\csname natexlab\endcsname\relax\def\natexlab#1{#1}\fi

\bibitem[{The()}]{TheC3}

\newblock \href {https://api.semanticscholar.org/CorpusID:268232499} {The claude 3 model family: Opus, sonnet, haiku}.

\bibitem[{Abdin et~al.(2024)Abdin, Aneja, Awadalla, Awadallah, Awan, Bach, Bahree, Bakhtiari, Bao, Behl, Benhaim, Bilenko, Bjorck, Bubeck, Cai, Cai, Chaudhary, Chen, Chen, Chen, Chen, Chen, Cheng, Chopra, Dai, Dixon, Eldan, Fragoso, Gao, Gao, Gao, Garg, Giorno, Goswami, Gunasekar, Haider, Hao, Hewett, Hu, Huynh, Iter, Jacobs, Javaheripi, Jin, Karampatziakis, Kauffmann, Khademi, Kim, Kim, Kurilenko, Lee, Lee, Li, Li, Liang, Liden, Lin, Lin, Liu, Liu, Liu, Liu, Liu, Luo, Madan, Mahmoudzadeh, Majercak, Mazzola, Mendes, Mitra, Modi, Nguyen, Norick, Patra, Perez-Becker, Portet, Pryzant, Qin, Radmilac, Ren, de~Rosa, Rosset, Roy, Ruwase, Saarikivi, Saied, Salim, Santacroce, Shah, Shang, Sharma, Shen, Shukla, Song, Tanaka, Tupini, Vaddamanu, Wang, Wang, Wang, Wang, Wang, Wang, Ward, Wen, Witte, Wu, Wu, Wyatt, Xiao, Xu, Xu, Xu, Xue, Yadav, Yang, Yang, Yang, Yang, Yu, Yuan, Zhang, Zhang, Zhang, Zhang, Zhang, Zhang, Zhang, and Zhou}]{abdin2024phi3technicalreporthighly}
Marah Abdin, Jyoti Aneja, Hany Awadalla, Ahmed Awadallah, Ammar~Ahmad Awan, Nguyen Bach, Amit Bahree, Arash Bakhtiari, Jianmin Bao, Harkirat Behl, Alon Benhaim, Misha Bilenko, Johan Bjorck, Sébastien Bubeck, Martin Cai, Qin Cai, Vishrav Chaudhary, Dong Chen, Dongdong Chen, Weizhu Chen, Yen-Chun Chen, Yi-Ling Chen, Hao Cheng, Parul Chopra, Xiyang Dai, Matthew Dixon, Ronen Eldan, Victor Fragoso, Jianfeng Gao, Mei Gao, Min Gao, Amit Garg, Allie~Del Giorno, Abhishek Goswami, Suriya Gunasekar, Emman Haider, Junheng Hao, Russell~J. Hewett, Wenxiang Hu, Jamie Huynh, Dan Iter, Sam~Ade Jacobs, Mojan Javaheripi, Xin Jin, Nikos Karampatziakis, Piero Kauffmann, Mahoud Khademi, Dongwoo Kim, Young~Jin Kim, Lev Kurilenko, James~R. Lee, Yin~Tat Lee, Yuanzhi Li, Yunsheng Li, Chen Liang, Lars Liden, Xihui Lin, Zeqi Lin, Ce~Liu, Liyuan Liu, Mengchen Liu, Weishung Liu, Xiaodong Liu, Chong Luo, Piyush Madan, Ali Mahmoudzadeh, David Majercak, Matt Mazzola, Caio César~Teodoro Mendes, Arindam Mitra, Hardik Modi, Anh Nguyen,
  Brandon Norick, Barun Patra, Daniel Perez-Becker, Thomas Portet, Reid Pryzant, Heyang Qin, Marko Radmilac, Liliang Ren, Gustavo de~Rosa, Corby Rosset, Sambudha Roy, Olatunji Ruwase, Olli Saarikivi, Amin Saied, Adil Salim, Michael Santacroce, Shital Shah, Ning Shang, Hiteshi Sharma, Yelong Shen, Swadheen Shukla, Xia Song, Masahiro Tanaka, Andrea Tupini, Praneetha Vaddamanu, Chunyu Wang, Guanhua Wang, Lijuan Wang, Shuohang Wang, Xin Wang, Yu~Wang, Rachel Ward, Wen Wen, Philipp Witte, Haiping Wu, Xiaoxia Wu, Michael Wyatt, Bin Xiao, Can Xu, Jiahang Xu, Weijian Xu, Jilong Xue, Sonali Yadav, Fan Yang, Jianwei Yang, Yifan Yang, Ziyi Yang, Donghan Yu, Lu~Yuan, Chenruidong Zhang, Cyril Zhang, Jianwen Zhang, Li~Lyna Zhang, Yi~Zhang, Yue Zhang, Yunan Zhang, and Xiren Zhou. 2024.
\newblock \href {http://arxiv.org/abs/2404.14219} {Phi-3 technical report: A highly capable language model locally on your phone}.

\bibitem[{Abraham and Lindström(2022)}]{abraham2022clevr}
Savitha~Sam Abraham and Adam~Dahlgren Lindström. 2022.
\newblock \href {https://arxiv.org/abs/2208.05358} {Clevr-math: A dataset for compositional language, visual and mathematical reasoning}.
\newblock \emph{arXiv preprint arXiv:2208.05358}.

\bibitem[{Agrawal et~al.(2024)Agrawal, Antoniak, Hanna, Bout, Chaplot, Chudnovsky, Costa, Monicault, Garg, Gervet, Ghosh, Héliou, Jacob, Jiang, Khandelwal, Lacroix, Lample, Casas, Lavril, Scao, Lo, Marshall, Martin, Mensch, Muddireddy, Nemychnikova, Pellat, Platen, Raghuraman, Rozière, Sablayrolles, Saulnier, Sauvestre, Shang, Soletskyi, Stewart, Stock, Studnia, Subramanian, Vaze, Wang, and Yang}]{agrawal2024pixtral12b}
Pravesh Agrawal, Szymon Antoniak, Emma~Bou Hanna, Baptiste Bout, Devendra Chaplot, Jessica Chudnovsky, Diogo Costa, Baudouin~De Monicault, Saurabh Garg, Theophile Gervet, Soham Ghosh, Amélie Héliou, Paul Jacob, Albert~Q. Jiang, Kartik Khandelwal, Timothée Lacroix, Guillaume Lample, Diego~Las Casas, Thibaut Lavril, Teven~Le Scao, Andy Lo, William Marshall, Louis Martin, Arthur Mensch, Pavankumar Muddireddy, Valera Nemychnikova, Marie Pellat, Patrick~Von Platen, Nikhil Raghuraman, Baptiste Rozière, Alexandre Sablayrolles, Lucile Saulnier, Romain Sauvestre, Wendy Shang, Roman Soletskyi, Lawrence Stewart, Pierre Stock, Joachim Studnia, Sandeep Subramanian, Sagar Vaze, Thomas Wang, and Sophia Yang. 2024.
\newblock \href {http://arxiv.org/abs/2410.07073} {Pixtral 12b}.

\bibitem[{Bai et~al.(2023)Bai, Bai, Yang, Wang, Tan, Wang, Lin, Zhou, and Zhou}]{bai2023qwenvl}
Jinze Bai, Shuai Bai, Shusheng Yang, Shijie Wang, Sinan Tan, Peng Wang, Junyang Lin, Chang Zhou, and Jingren Zhou. 2023.
\newblock \href {http://arxiv.org/abs/2308.12966} {Qwen-vl: A versatile vision-language model for understanding, localization, text reading, and beyond}.

\bibitem[{Bai et~al.(2022)Bai, Jones, Ndousse, Askell, Chen, DasSarma, Drain, Fort, Ganguli, Henighan et~al.}]{bai2022training}
Yuntao Bai, Andy Jones, Kamal Ndousse, Amanda Askell, Anna Chen, Nova DasSarma, Dawn Drain, Stanislav Fort, Deep Ganguli, Tom Henighan, et~al. 2022.
\newblock Training a helpful and harmless assistant with reinforcement learning from human feedback.
\newblock \emph{arXiv preprint arXiv:2204.05862}.

\bibitem[{Brown et~al.(2020)Brown, Mann, Ryder, Subbiah, Kaplan, Dhariwal, Neelakantan, Shyam, Sastry, Askell et~al.}]{brown2020language}
Tom Brown, Benjamin Mann, Nick Ryder, Melanie Subbiah, Jared~D Kaplan, Prafulla Dhariwal, Arvind Neelakantan, Pranav Shyam, Girish Sastry, Amanda Askell, et~al. 2020.
\newblock Language models are few-shot learners.
\newblock \emph{Advances in neural information processing systems}, 33:1877--1901.

\bibitem[{Chen et~al.(2024{\natexlab{a}})Chen, Chen, Zhang, Liu, Wang, Zhou, Zhang, Wan, Zhou, and Sun}]{chen2024mllmasajudgeassessingmultimodalllmasajudge}
Dongping Chen, Ruoxi Chen, Shilin Zhang, Yinuo Liu, Yaochen Wang, Huichi Zhou, Qihui Zhang, Yao Wan, Pan Zhou, and Lichao Sun. 2024{\natexlab{a}}.
\newblock \href {http://arxiv.org/abs/2402.04788} {Mllm-as-a-judge: Assessing multimodal llm-as-a-judge with vision-language benchmark}.

\bibitem[{Chen et~al.(2024{\natexlab{b}})Chen, Cao, Zhang, and Lu}]{chen2024quantifyingmitigatingunimodalbiases}
Meiqi Chen, Yixin Cao, Yan Zhang, and Chaochao Lu. 2024{\natexlab{b}}.
\newblock \href {http://arxiv.org/abs/2403.18346} {Quantifying and mitigating unimodal biases in multimodal large language models: A causal perspective}.

\bibitem[{Chen et~al.(2024{\natexlab{c}})Chen, Wu, Wang, Su, Chen, Xing, Zhong, Zhang, Zhu, Lu et~al.}]{chen2024internvl}
Zhe Chen, Jiannan Wu, Wenhai Wang, Weijie Su, Guo Chen, Sen Xing, Muyan Zhong, Qinglong Zhang, Xizhou Zhu, Lewei Lu, et~al. 2024{\natexlab{c}}.
\newblock Internvl: Scaling up vision foundation models and aligning for generic visual-linguistic tasks.
\newblock In \emph{Proceedings of the IEEE/CVF Conference on Computer Vision and Pattern Recognition}, pages 24185--24198.

\bibitem[{Chowdhery et~al.(2022)Chowdhery, Narang, Devlin, Bosma, Mishra, Roberts, Barham, Chung, Sutton, Gehrmann et~al.}]{chowdhery2022palm}
Aakanksha Chowdhery, Sharan Narang, Jacob Devlin, Maarten Bosma, Gaurav Mishra, Adam Roberts, Paul Barham, Hyung~Won Chung, Charles Sutton, Sebastian Gehrmann, et~al. 2022.
\newblock Palm: Scaling language modeling with pathways.
\newblock \emph{arXiv preprint arXiv:2204.02311}.

\bibitem[{Dai et~al.(2025)Dai, Sun, Zhao, Liu, Li, Gao, and Li}]{dai2025carevl}
Muzhi Dai, Jiashuo Sun, Zhiyuan Zhao, Shixuan Liu, Rui Li, Junyu Gao, and Xuelong Li. 2025.
\newblock From captions to rewards (carevl): Leveraging large language model experts for enhanced reward modeling in large vision-language models.
\newblock \emph{arXiv preprint arXiv:2503.06260}.

\bibitem[{Dai et~al.(2024)Dai, Lee, Wang, Yang, Liu, Barker, Rintamaki, Shoeybi, Catanzaro, and Ping}]{dai2024nvlmopenfrontierclassmultimodal}
Wenliang Dai, Nayeon Lee, Boxin Wang, Zhuolin Yang, Zihan Liu, Jon Barker, Tuomas Rintamaki, Mohammad Shoeybi, Bryan Catanzaro, and Wei Ping. 2024.
\newblock \href {http://arxiv.org/abs/2409.11402} {Nvlm: Open frontier-class multimodal llms}.

\bibitem[{Dai et~al.(2023)Dai, Li, Li, Tiong, Zhao, Wang, Li, Fung, and Hoi}]{dai2023instructblip}
Wenliang Dai, Junnan Li, Dongxu Li, Anthony Meng~Huat Tiong, Junqi Zhao, Weisheng Wang, Boyang Li, Pascale Fung, and Steven Hoi. 2023.
\newblock \href {http://arxiv.org/abs/2305.06500} {Instructblip: Towards general-purpose vision-language models with instruction tuning}.

\bibitem[{DeepSeek-AI et~al.(2025)DeepSeek-AI, Guo, Yang, Zhang, Song, Zhang, Xu, Zhu, Ma, Wang, Bi, Zhang, Yu, Wu, Wu, Gou, Shao, Li, Gao, Liu, Xue, Wang, Wu, Feng, Lu, Zhao, Deng, Zhang, Ruan, Dai, Chen, Ji, Li, Lin, Dai, Luo, Hao, Chen, Li, Zhang, Bao, Xu, Wang, Ding, Xin, Gao, Qu, Li, Guo, Li, Wang, Chen, Yuan, Qiu, Li, Cai, Ni, Liang, Chen, Dong, Hu, Gao, Guan, Huang, Yu, Wang, Zhang, Zhao, Wang, Zhang, Xu, Xia, Zhang, Zhang, Tang, Li, Wang, Li, Tian, Huang, Zhang, Wang, Chen, Du, Ge, Zhang, Pan, Wang, Chen, Jin, Chen, Lu, Zhou, Chen, Ye, Wang, Yu, Zhou, Pan, Li, Zhou, Wu, Ye, Yun, Pei, Sun, Wang, Zeng, Zhao, Liu, Liang, Gao, Yu, Zhang, Xiao, An, Liu, Wang, Chen, Nie, Cheng, Liu, Xie, Liu, Yang, Li, Su, Lin, Li, Jin, Shen, Chen, Sun, Wang, Song, Zhou, Wang, Shan, Li, Wang, Wei, Zhang, Xu, Li, Zhao, Sun, Wang, Yu, Zhang, Shi, Xiong, He, Piao, Wang, Tan, Ma, Liu, Guo, Ou, Wang, Gong, Zou, He, Xiong, Luo, You, Liu, Zhou, Zhu, Xu, Huang, Li, Zheng, Zhu, Ma, Tang, Zha, Yan, Ren, Ren, Sha, Fu, Xu, Xie, Zhang,
  Hao, Ma, Yan, Wu, Gu, Zhu, Liu, Li, Xie, Song, Pan, Huang, Xu, Zhang, and Zhang}]{deepseekai2025deepseekr1incentivizingreasoningcapability}
DeepSeek-AI, Daya Guo, Dejian Yang, Haowei Zhang, Junxiao Song, Ruoyu Zhang, Runxin Xu, Qihao Zhu, Shirong Ma, Peiyi Wang, Xiao Bi, Xiaokang Zhang, Xingkai Yu, Yu~Wu, Z.~F. Wu, Zhibin Gou, Zhihong Shao, Zhuoshu Li, Ziyi Gao, Aixin Liu, Bing Xue, Bingxuan Wang, Bochao Wu, Bei Feng, Chengda Lu, Chenggang Zhao, Chengqi Deng, Chenyu Zhang, Chong Ruan, Damai Dai, Deli Chen, Dongjie Ji, Erhang Li, Fangyun Lin, Fucong Dai, Fuli Luo, Guangbo Hao, Guanting Chen, Guowei Li, H.~Zhang, Han Bao, Hanwei Xu, Haocheng Wang, Honghui Ding, Huajian Xin, Huazuo Gao, Hui Qu, Hui Li, Jianzhong Guo, Jiashi Li, Jiawei Wang, Jingchang Chen, Jingyang Yuan, Junjie Qiu, Junlong Li, J.~L. Cai, Jiaqi Ni, Jian Liang, Jin Chen, Kai Dong, Kai Hu, Kaige Gao, Kang Guan, Kexin Huang, Kuai Yu, Lean Wang, Lecong Zhang, Liang Zhao, Litong Wang, Liyue Zhang, Lei Xu, Leyi Xia, Mingchuan Zhang, Minghua Zhang, Minghui Tang, Meng Li, Miaojun Wang, Mingming Li, Ning Tian, Panpan Huang, Peng Zhang, Qiancheng Wang, Qinyu Chen, Qiushi Du, Ruiqi Ge, Ruisong
  Zhang, Ruizhe Pan, Runji Wang, R.~J. Chen, R.~L. Jin, Ruyi Chen, Shanghao Lu, Shangyan Zhou, Shanhuang Chen, Shengfeng Ye, Shiyu Wang, Shuiping Yu, Shunfeng Zhou, Shuting Pan, S.~S. Li, Shuang Zhou, Shaoqing Wu, Shengfeng Ye, Tao Yun, Tian Pei, Tianyu Sun, T.~Wang, Wangding Zeng, Wanjia Zhao, Wen Liu, Wenfeng Liang, Wenjun Gao, Wenqin Yu, Wentao Zhang, W.~L. Xiao, Wei An, Xiaodong Liu, Xiaohan Wang, Xiaokang Chen, Xiaotao Nie, Xin Cheng, Xin Liu, Xin Xie, Xingchao Liu, Xinyu Yang, Xinyuan Li, Xuecheng Su, Xuheng Lin, X.~Q. Li, Xiangyue Jin, Xiaojin Shen, Xiaosha Chen, Xiaowen Sun, Xiaoxiang Wang, Xinnan Song, Xinyi Zhou, Xianzu Wang, Xinxia Shan, Y.~K. Li, Y.~Q. Wang, Y.~X. Wei, Yang Zhang, Yanhong Xu, Yao Li, Yao Zhao, Yaofeng Sun, Yaohui Wang, Yi~Yu, Yichao Zhang, Yifan Shi, Yiliang Xiong, Ying He, Yishi Piao, Yisong Wang, Yixuan Tan, Yiyang Ma, Yiyuan Liu, Yongqiang Guo, Yuan Ou, Yuduan Wang, Yue Gong, Yuheng Zou, Yujia He, Yunfan Xiong, Yuxiang Luo, Yuxiang You, Yuxuan Liu, Yuyang Zhou, Y.~X. Zhu,
  Yanhong Xu, Yanping Huang, Yaohui Li, Yi~Zheng, Yuchen Zhu, Yunxian Ma, Ying Tang, Yukun Zha, Yuting Yan, Z.~Z. Ren, Zehui Ren, Zhangli Sha, Zhe Fu, Zhean Xu, Zhenda Xie, Zhengyan Zhang, Zhewen Hao, Zhicheng Ma, Zhigang Yan, Zhiyu Wu, Zihui Gu, Zijia Zhu, Zijun Liu, Zilin Li, Ziwei Xie, Ziyang Song, Zizheng Pan, Zhen Huang, Zhipeng Xu, Zhongyu Zhang, and Zhen Zhang. 2025.
\newblock \href {http://arxiv.org/abs/2501.12948} {Deepseek-r1: Incentivizing reasoning capability in llms via reinforcement learning}.

\bibitem[{Deitke et~al.(2024)Deitke, Clark, Lee, Tripathi, Yang, Park, Salehi, Muennighoff, Lo, Soldaini, Lu, Anderson, Bransom, Ehsani, Ngo, Chen, Patel, Yatskar, Callison-Burch, Head, Hendrix, Bastani, VanderBilt, Lambert, Chou, Chheda, Sparks, Skjonsberg, Schmitz, Sarnat, Bischoff, Walsh, Newell, Wolters, Gupta, Zeng, Borchardt, Groeneveld, Nam, Lebrecht, Wittlif, Schoenick, Michel, Krishna, Weihs, Smith, Hajishirzi, Girshick, Farhadi, and Kembhavi}]{deitke2024molmopixmoopenweights}
Matt Deitke, Christopher Clark, Sangho Lee, Rohun Tripathi, Yue Yang, Jae~Sung Park, Mohammadreza Salehi, Niklas Muennighoff, Kyle Lo, Luca Soldaini, Jiasen Lu, Taira Anderson, Erin Bransom, Kiana Ehsani, Huong Ngo, YenSung Chen, Ajay Patel, Mark Yatskar, Chris Callison-Burch, Andrew Head, Rose Hendrix, Favyen Bastani, Eli VanderBilt, Nathan Lambert, Yvonne Chou, Arnavi Chheda, Jenna Sparks, Sam Skjonsberg, Michael Schmitz, Aaron Sarnat, Byron Bischoff, Pete Walsh, Chris Newell, Piper Wolters, Tanmay Gupta, Kuo-Hao Zeng, Jon Borchardt, Dirk Groeneveld, Crystal Nam, Sophie Lebrecht, Caitlin Wittlif, Carissa Schoenick, Oscar Michel, Ranjay Krishna, Luca Weihs, Noah~A. Smith, Hannaneh Hajishirzi, Ross Girshick, Ali Farhadi, and Aniruddha Kembhavi. 2024.
\newblock \href {http://arxiv.org/abs/2409.17146} {Molmo and pixmo: Open weights and open data for state-of-the-art vision-language models}.

\bibitem[{Dong et~al.(2023)}]{dong2023unified}
Junjie Dong et~al. 2023.
\newblock Unified image reward modeling and learning from human feedback.
\newblock \emph{arXiv preprint arXiv:2306.05645}.

\bibitem[{for AI(2019)}]{ai2d}
Allen~Institute for AI. 2019.
\newblock Ai2 diagram dataset (ai2d).
\newblock \url{https://registry.opendata.aws/allenai-diagrams/}.
\newblock Accessed: 2025-05-08.

\bibitem[{Gao et~al.(2023{\natexlab{a}})Gao, Pi, Zhang, Ye, Zhong, Wang, Hong, Han, Xu, Li, and Kong}]{gao2023gllava}
Jiahui Gao, Renjie Pi, Jipeng Zhang, Jiacheng Ye, Wanjun Zhong, Yufei Wang, Lanqing Hong, Jianhua Han, Hang Xu, Zhenguo Li, and Lingpeng Kong. 2023{\natexlab{a}}.
\newblock \href {http://arxiv.org/abs/2312.11370} {G-llava: Solving geometric problem with multi-modal large language model}.

\bibitem[{Gao et~al.(2023{\natexlab{b}})Gao, Han, Zhang, Lin, Geng, Zhou, Zhang, Lu, He, Yue, Li, and Qiao}]{gao2023llamaadapter}
Peng Gao, Jiaming Han, Renrui Zhang, Ziyi Lin, Shijie Geng, Aojun Zhou, Wei Zhang, Pan Lu, Conghui He, Xiangyu Yue, Hongsheng Li, and Yu~Qiao. 2023{\natexlab{b}}.
\newblock \href {http://arxiv.org/abs/2304.15010} {Llama-adapter v2: Parameter-efficient visual instruction model}.

\bibitem[{{GEOS contributors}(2024)}]{geos2024}
{GEOS contributors}. 2024.
\newblock \href {https://doi.org/10.5281/zenodo.11396894} {\emph{GEOS Computational Geometry Library}}.
\newblock Open Source Geospatial Foundation.

\bibitem[{Guo et~al.(2024)Guo, Zhang, Liu, Liu, Khalman, Llinares, Rame, Mesnard, Zhao, Piot, Ferret, and Blondel}]{guo2024directlanguagemodelalignment}
Shangmin Guo, Biao Zhang, Tianlin Liu, Tianqi Liu, Misha Khalman, Felipe Llinares, Alexandre Rame, Thomas Mesnard, Yao Zhao, Bilal Piot, Johan Ferret, and Mathieu Blondel. 2024.
\newblock \href {http://arxiv.org/abs/2402.04792} {Direct language model alignment from online ai feedback}.

\bibitem[{Hoffmann et~al.(2022)Hoffmann, Borgeaud, Mensch, Buchatskaya, Cai, Rutherford, Casas, Hendricks, Welbl, Clark et~al.}]{hoffmann2022training}
Jordan Hoffmann, Sebastian Borgeaud, Arthur Mensch, Elena Buchatskaya, Trevor Cai, Eliza Rutherford, Diego de~Las Casas, Lisa~Anne Hendricks, Johannes Welbl, Aidan Clark, et~al. 2022.
\newblock Training compute-optimal large language models.
\newblock \emph{arXiv preprint arXiv:2203.15556}.

\bibitem[{Hosseini et~al.(2024)Hosseini, Yuan, Malkin, Courville, Sordoni, and Agarwal}]{hosseini2024vstartrainingverifiersselftaught}
Arian Hosseini, Xingdi Yuan, Nikolay Malkin, Aaron Courville, Alessandro Sordoni, and Rishabh Agarwal. 2024.
\newblock \href {http://arxiv.org/abs/2402.06457} {V-star: Training verifiers for self-taught reasoners}.

\bibitem[{Jiang et~al.(2023)Jiang, Ren, and Lin}]{jiang2023llmblenderensemblinglargelanguage}
Dongfu Jiang, Xiang Ren, and Bill~Yuchen Lin. 2023.
\newblock \href {http://arxiv.org/abs/2306.02561} {Llm-blender: Ensembling large language models with pairwise ranking and generative fusion}.

\bibitem[{Johnson et~al.(2017)Johnson, Hariharan, van~der Maaten, Fei-Fei, Zitnick, and Girshick}]{johnson2017clevr}
Justin Johnson, Bharath Hariharan, Laurens van~der Maaten, Li~Fei-Fei, C~Lawrence Zitnick, and Ross Girshick. 2017.
\newblock \href {https://arxiv.org/abs/1612.06890} {Clevr: A diagnostic dataset for compositional language and elementary visual reasoning}.
\newblock In \emph{Proceedings of the IEEE Conference on Computer Vision and Pattern Recognition (CVPR)}, pages 2901--2910.

\bibitem[{Kafle et~al.(2018)Kafle, Price, Cohen, and Kanan}]{kafle2018dvqa}
Kushal Kafle, Brian Price, Scott Cohen, and Christopher Kanan. 2018.
\newblock \href {https://arxiv.org/abs/1801.08163} {Dvqa: Understanding data visualizations via question answering}.
\newblock In \emph{Proceedings of the IEEE Conference on Computer Vision and Pattern Recognition (CVPR)}, pages 5515--5524.

\bibitem[{Kahou et~al.(2018)Kahou, Michalski, Atkinson, Kádár, Trischler, and Bengio}]{kahou2018figureqa}
Samira~Ebrahimi Kahou, Vincent Michalski, Adam Atkinson, Ákos Kádár, Adam Trischler, and Yoshua Bengio. 2018.
\newblock \href {https://openreview.net/forum?id=H1mz0OyDz} {Figureqa: An annotated figure dataset for visual reasoning}.
\newblock In \emph{Proceedings of the International Conference on Learning Representations (ICLR) Workshops}.

\bibitem[{Kantharaj et~al.(2022)Kantharaj, Leong, Lin, Masry, Thakkar, Hoque, and Joty}]{kantharaj2022chart}
Shankar Kantharaj, Rixie Tiffany~Ko Leong, Xiang Lin, Ahmed Masry, Megh Thakkar, Enamul Hoque, and Shafiq Joty. 2022.
\newblock \href {https://aclanthology.org/2022.acl-long.277/} {Chart-to-text: A large-scale benchmark for chart summarization}.
\newblock In \emph{Proceedings of the 60th Annual Meeting of the Association for Computational Linguistics (Volume 1: Long Papers)}, pages 3987--3998.

\bibitem[{Kumar et~al.(2024)Kumar, Zhuang, Agarwal, Su, Co-Reyes, Singh, Baumli, Iqbal, Bishop, Roelofs, Zhang, McKinney, Shrivastava, Paduraru, Tucker, Precup, Behbahani, and Faust}]{kumar2024traininglanguagemodelsselfcorrect}
Aviral Kumar, Vincent Zhuang, Rishabh Agarwal, Yi~Su, John~D Co-Reyes, Avi Singh, Kate Baumli, Shariq Iqbal, Colton Bishop, Rebecca Roelofs, Lei~M Zhang, Kay McKinney, Disha Shrivastava, Cosmin Paduraru, George Tucker, Doina Precup, Feryal Behbahani, and Aleksandra Faust. 2024.
\newblock \href {http://arxiv.org/abs/2409.12917} {Training language models to self-correct via reinforcement learning}.

\bibitem[{Lee et~al.(2023)Lee, Saharia, Li, Ho, Chen, Zhang, Li, Hovy, Salakhutdinov, Le et~al.}]{lee2023image}
Kevin Lee, Chitwan Saharia, Xi~Li, Jonathan Ho, Daniel Chen, Ruoming Zhang, Xuezhi Li, Eduard Hovy, Ruslan Salakhutdinov, Quoc Le, et~al. 2023.
\newblock Image reward models for reinforcement learning with human feedback.
\newblock In \emph{Proceedings of the IEEE/CVF Conference on Computer Vision and Pattern Recognition}, pages 5001--5010.

\bibitem[{Li et~al.(2024{\natexlab{a}})Li, Zhang, Guo, Zhang, Li, Zhang, Zhang, Li, Liu, and Li}]{li2024llavaonevisioneasyvisualtask}
Bo~Li, Yuanhan Zhang, Dong Guo, Renrui Zhang, Feng Li, Hao Zhang, Kaichen Zhang, Yanwei Li, Ziwei Liu, and Chunyuan Li. 2024{\natexlab{a}}.
\newblock \href {http://arxiv.org/abs/2408.03326} {Llava-onevision: Easy visual task transfer}.

\bibitem[{Li et~al.(2023{\natexlab{a}})Li, Li, Savarese, and Hoi}]{li2023blip2}
Junnan Li, Dongxu Li, Silvio Savarese, and Steven Hoi. 2023{\natexlab{a}}.
\newblock \href {http://arxiv.org/abs/2301.12597} {Blip-2: Bootstrapping language-image pre-training with frozen image encoders and large language models}.

\bibitem[{Li et~al.(2024{\natexlab{b}})Li, Wei, Xie, Yang, Song, Wang, An, Liu, Li, Lin, Kong, and Liu}]{li2024vlrewardbenchchallengingbenchmarkvisionlanguage}
Lei Li, Yuancheng Wei, Zhihui Xie, Xuqing Yang, Yifan Song, Peiyi Wang, Chenxin An, Tianyu Liu, Sujian Li, Bill~Yuchen Lin, Lingpeng Kong, and Qi~Liu. 2024{\natexlab{b}}.
\newblock \href {http://arxiv.org/abs/2411.17451} {Vlrewardbench: A challenging benchmark for vision-language generative reward models}.

\bibitem[{Li et~al.(2023{\natexlab{b}})Li, Xie, Li, Chen, Wang, Chen, Yang, Wang, and Kong}]{li2023silkie}
Lei Li, Zhihui Xie, Mukai Li, Shunian Chen, Peiyi Wang, Liang Chen, Yazheng Yang, Benyou Wang, and Lingpeng Kong. 2023{\natexlab{b}}.
\newblock \href {http://arxiv.org/abs/2312.10665} {Silkie: Preference distillation for large visual language models}.

\bibitem[{Li et~al.(2023{\natexlab{c}})Li, Yang, and Wang}]{li2023reinforcementlearninghumanfeedback}
Zihao Li, Zhuoran Yang, and Mengdi Wang. 2023{\natexlab{c}}.
\newblock \href {http://arxiv.org/abs/2305.18438} {Reinforcement learning with human feedback: Learning dynamic choices via pessimism}.

\bibitem[{Liu et~al.(2024)Liu, Zeng, Liu, Yan, He, Wang, Yan, Liu, and Zhou}]{liu2024skyworkrewardbagtricksreward}
Chris~Yuhao Liu, Liang Zeng, Jiacai Liu, Rui Yan, Jujie He, Chaojie Wang, Shuicheng Yan, Yang Liu, and Yahui Zhou. 2024.
\newblock \href {http://arxiv.org/abs/2410.18451} {Skywork-reward: Bag of tricks for reward modeling in llms}.

\bibitem[{Liu et~al.(2023)Liu, Li, Wu, and Lee}]{liu2023llava}
Haotian Liu, Chunyuan Li, Qingyang Wu, and Yong~Jae Lee. 2023.
\newblock Visual instruction tuning.

\bibitem[{Liu et~al.(2025)Liu, Wang, Xu, Ma, Ruan, Li, Liu, and Wu}]{liu2025inferencetimescalinggeneralistreward}
Zijun Liu, Peiyi Wang, Runxin Xu, Shirong Ma, Chong Ruan, Peng Li, Yang Liu, and Yu~Wu. 2025.
\newblock \href {http://arxiv.org/abs/2504.02495} {Inference-time scaling for generalist reward modeling}.

\bibitem[{Lu et~al.(2024)Lu, Liu, Zhang, Wang, Dong, Liu, Sun, Ren, Li, Sun et~al.}]{lu2024deepseek}
Haoyu Lu, Wen Liu, Bo~Zhang, Bingxuan Wang, Kai Dong, Bo~Liu, Jingxiang Sun, Tongzheng Ren, Zhuoshu Li, Yaofeng Sun, et~al. 2024.
\newblock Deepseek-vl: towards real-world vision-language understanding.
\newblock \emph{arXiv preprint arXiv:2403.05525}.

\bibitem[{Lu et~al.(2022)Lu, Mishra, Xia, Qiu, Chang, Zhu, Tafjord, Clark, and Kalyan}]{lu2022learn}
Pan Lu, Swaroop Mishra, Tony Xia, Liang Qiu, Kai-Wei Chang, Song-Chun Zhu, Oyvind Tafjord, Peter Clark, and Ashwin Kalyan. 2022.
\newblock \href {https://arxiv.org/abs/2209.09513} {Learn to explain: Multimodal reasoning via thought chains for science question answering}.
\newblock In \emph{Advances in Neural Information Processing Systems}.

\bibitem[{Masry et~al.(2022)Masry, Long, Tan, Joty, and Hoque}]{masry2022chartqa}
Ahmed Masry, Do~Xuan Long, Jia~Qing Tan, Shafiq Joty, and Enamul Hoque. 2022.
\newblock \href {https://aclanthology.org/2022.findings-acl.177/} {Chartqa: A benchmark for question answering about charts with visual and logical reasoning}.
\newblock In \emph{Findings of the Association for Computational Linguistics: ACL 2022}, pages 2225--2239.

\bibitem[{Meng et~al.(2025)Meng, Du, Liu, Zhou, Lu, Fu, Han, Shi, Wang, He, Zhang, Luo, Qiao, Zhang, and Shao}]{meng2025mmeurekaexploringfrontiersmultimodal}
Fanqing Meng, Lingxiao Du, Zongkai Liu, Zhixiang Zhou, Quanfeng Lu, Daocheng Fu, Tiancheng Han, Botian Shi, Wenhai Wang, Junjun He, Kaipeng Zhang, Ping Luo, Yu~Qiao, Qiaosheng Zhang, and Wenqi Shao. 2025.
\newblock \href {http://arxiv.org/abs/2503.07365} {Mm-eureka: Exploring the frontiers of multimodal reasoning with rule-based reinforcement learning}.

\bibitem[{OpenAI et~al.(2024{\natexlab{a}})OpenAI, :, Hurst, Lerer, Goucher, Perelman, Ramesh, Clark, Ostrow, Welihinda, Hayes, Radford, Mądry, Baker-Whitcomb, Beutel, Borzunov, Carney, Chow, Kirillov, Nichol, Paino, Renzin, Passos, Kirillov, Christakis, Conneau, Kamali, Jabri, Moyer, Tam, Crookes, Tootoochian, Tootoonchian, Kumar, Vallone, Karpathy, Braunstein, Cann, Codispoti, Galu, Kondrich, Tulloch, Mishchenko, Baek, Jiang, Pelisse, Woodford, Gosalia, Dhar, Pantuliano, Nayak, Oliver, Zoph, Ghorbani, Leimberger, Rossen, Sokolowsky, Wang, Zweig, Hoover, Samic, McGrew, Spero, Giertler, Cheng, Lightcap, Walkin, Quinn, Guarraci, Hsu, Kellogg, Eastman, Lugaresi, Wainwright, Bassin, Hudson, Chu, Nelson, Li, Shern, Conger, Barette, Voss, Ding, Lu, Zhang, Beaumont, Hallacy, Koch, Gibson, Kim, Choi, McLeavey, Hesse, Fischer, Winter, Czarnecki, Jarvis, Wei, Koumouzelis, Sherburn, Kappler, Levin, Levy, Carr, Farhi, Mely, Robinson, Sasaki, Jin, Valladares, Tsipras, Li, Nguyen, Findlay, Oiwoh, Wong, Asdar, Proehl,
  Yang, Antonow, Kramer, Peterson, Sigler, Wallace, Brevdo, Mays, Khorasani, Such, Raso, Zhang, von Lohmann, Sulit, Goh, Oden, Salmon, Starace, Brockman, Salman, Bao, Hu, Wong, Wang, Schmidt, Whitney, Jun, Kirchner, de~Oliveira~Pinto, Ren, Chang, Chung, Kivlichan, O'Connell, O'Connell, Osband, Silber, Sohl, Okuyucu, Lan, Kostrikov, Sutskever, Kanitscheider, Gulrajani, Coxon, Menick, Pachocki, Aung, Betker, Crooks, Lennon, Kiros, Leike, Park, Kwon, Phang, Teplitz, Wei, Wolfe, Chen, Harris, Varavva, Lee, Shieh, Lin, Yu, Weng, Tang, Yu, Jang, Candela, Beutler, Landers, Parish, Heidecke, Schulman, Lachman, McKay, Uesato, Ward, Kim, Huizinga, Sitkin, Kraaijeveld, Gross, Kaplan, Snyder, Achiam, Jiao, Lee, Zhuang, Harriman, Fricke, Hayashi, Singhal, Shi, Karthik, Wood, Rimbach, Hsu, Nguyen, Gu-Lemberg, Button, Liu, Howe, Muthukumar, Luther, Ahmad, Kai, Itow, Workman, Pathak, Chen, Jing, Guy, Fedus, Zhou, Mamitsuka, Weng, McCallum, Held, Ouyang, Feuvrier, Zhang, Kondraciuk, Kaiser, Hewitt, Metz, Doshi, Aflak, Simens,
  Boyd, Thompson, Dukhan, Chen, Gray, Hudnall, Zhang, Aljubeh, Litwin, Zeng, Johnson, Shetty, Gupta, Shah, Yatbaz, Yang, Zhong, Glaese, Chen, Janner, Lampe, Petrov, Wu, Wang, Fradin, Pokrass, Castro, de~Castro, Pavlov, Brundage, Wang, Khan, Murati, Bavarian, Lin, Yesildal, Soto, Gimelshein, Cone, Staudacher, Summers, LaFontaine, Chowdhury, Ryder, Stathas, Turley, Tezak, Felix, Kudige, Keskar, Deutsch, Bundick, Puckett, Nachum, Okelola, Boiko, Murk, Jaffe, Watkins, Godement, Campbell-Moore, Chao, McMillan, Belov, Su, Bak, Bakkum, Deng, Dolan, Hoeschele, Welinder, Tillet, Pronin, Tillet, Dhariwal, Yuan, Dias, Lim, Arora, Troll, Lin, Lopes, Puri, Miyara, Leike, Gaubert, Zamani, Wang, Donnelly, Honsby, Smith, Sahai, Ramchandani, Huet, Carmichael, Zellers, Chen, Chen, Nigmatullin, Cheu, Jain, Altman, Schoenholz, Toizer, Miserendino, Agarwal, Culver, Ethersmith, Gray, Grove, Metzger, Hermani, Jain, Zhao, Wu, Jomoto, Wu, Shuaiqi, Xia, Phene, Papay, Narayanan, Coffey, Lee, Hall, Balaji, Broda, Stramer, Xu, Gogineni,
  Christianson, Sanders, Patwardhan, Cunninghman, Degry, Dimson, Raoux, Shadwell, Zheng, Underwood, Markov, Sherbakov, Rubin, Stasi, Kaftan, Heywood, Peterson, Walters, Eloundou, Qi, Moeller, Monaco, Kuo, Fomenko, Chang, Zheng, Zhou, Manassra, Sheu, Zaremba, Patil, Qian, Kim, Cheng, Zhang, He, Zhang, Jin, Dai, and Malkov}]{openai2024gpt4ocard}
OpenAI, :, Aaron Hurst, Adam Lerer, Adam~P. Goucher, Adam Perelman, Aditya Ramesh, Aidan Clark, AJ~Ostrow, Akila Welihinda, Alan Hayes, Alec Radford, Aleksander Mądry, Alex Baker-Whitcomb, Alex Beutel, Alex Borzunov, Alex Carney, Alex Chow, Alex Kirillov, Alex Nichol, Alex Paino, Alex Renzin, Alex~Tachard Passos, Alexander Kirillov, Alexi Christakis, Alexis Conneau, Ali Kamali, Allan Jabri, Allison Moyer, Allison Tam, Amadou Crookes, Amin Tootoochian, Amin Tootoonchian, Ananya Kumar, Andrea Vallone, Andrej Karpathy, Andrew Braunstein, Andrew Cann, Andrew Codispoti, Andrew Galu, Andrew Kondrich, Andrew Tulloch, Andrey Mishchenko, Angela Baek, Angela Jiang, Antoine Pelisse, Antonia Woodford, Anuj Gosalia, Arka Dhar, Ashley Pantuliano, Avi Nayak, Avital Oliver, Barret Zoph, Behrooz Ghorbani, Ben Leimberger, Ben Rossen, Ben Sokolowsky, Ben Wang, Benjamin Zweig, Beth Hoover, Blake Samic, Bob McGrew, Bobby Spero, Bogo Giertler, Bowen Cheng, Brad Lightcap, Brandon Walkin, Brendan Quinn, Brian Guarraci, Brian Hsu,
  Bright Kellogg, Brydon Eastman, Camillo Lugaresi, Carroll Wainwright, Cary Bassin, Cary Hudson, Casey Chu, Chad Nelson, Chak Li, Chan~Jun Shern, Channing Conger, Charlotte Barette, Chelsea Voss, Chen Ding, Cheng Lu, Chong Zhang, Chris Beaumont, Chris Hallacy, Chris Koch, Christian Gibson, Christina Kim, Christine Choi, Christine McLeavey, Christopher Hesse, Claudia Fischer, Clemens Winter, Coley Czarnecki, Colin Jarvis, Colin Wei, Constantin Koumouzelis, Dane Sherburn, Daniel Kappler, Daniel Levin, Daniel Levy, David Carr, David Farhi, David Mely, David Robinson, David Sasaki, Denny Jin, Dev Valladares, Dimitris Tsipras, Doug Li, Duc~Phong Nguyen, Duncan Findlay, Edede Oiwoh, Edmund Wong, Ehsan Asdar, Elizabeth Proehl, Elizabeth Yang, Eric Antonow, Eric Kramer, Eric Peterson, Eric Sigler, Eric Wallace, Eugene Brevdo, Evan Mays, Farzad Khorasani, Felipe~Petroski Such, Filippo Raso, Francis Zhang, Fred von Lohmann, Freddie Sulit, Gabriel Goh, Gene Oden, Geoff Salmon, Giulio Starace, Greg Brockman, Hadi
  Salman, Haiming Bao, Haitang Hu, Hannah Wong, Haoyu Wang, Heather Schmidt, Heather Whitney, Heewoo Jun, Hendrik Kirchner, Henrique~Ponde de~Oliveira~Pinto, Hongyu Ren, Huiwen Chang, Hyung~Won Chung, Ian Kivlichan, Ian O'Connell, Ian O'Connell, Ian Osband, Ian Silber, Ian Sohl, Ibrahim Okuyucu, Ikai Lan, Ilya Kostrikov, Ilya Sutskever, Ingmar Kanitscheider, Ishaan Gulrajani, Jacob Coxon, Jacob Menick, Jakub Pachocki, James Aung, James Betker, James Crooks, James Lennon, Jamie Kiros, Jan Leike, Jane Park, Jason Kwon, Jason Phang, Jason Teplitz, Jason Wei, Jason Wolfe, Jay Chen, Jeff Harris, Jenia Varavva, Jessica~Gan Lee, Jessica Shieh, Ji~Lin, Jiahui Yu, Jiayi Weng, Jie Tang, Jieqi Yu, Joanne Jang, Joaquin~Quinonero Candela, Joe Beutler, Joe Landers, Joel Parish, Johannes Heidecke, John Schulman, Jonathan Lachman, Jonathan McKay, Jonathan Uesato, Jonathan Ward, Jong~Wook Kim, Joost Huizinga, Jordan Sitkin, Jos Kraaijeveld, Josh Gross, Josh Kaplan, Josh Snyder, Joshua Achiam, Joy Jiao, Joyce Lee, Juntang
  Zhuang, Justyn Harriman, Kai Fricke, Kai Hayashi, Karan Singhal, Katy Shi, Kavin Karthik, Kayla Wood, Kendra Rimbach, Kenny Hsu, Kenny Nguyen, Keren Gu-Lemberg, Kevin Button, Kevin Liu, Kiel Howe, Krithika Muthukumar, Kyle Luther, Lama Ahmad, Larry Kai, Lauren Itow, Lauren Workman, Leher Pathak, Leo Chen, Li~Jing, Lia Guy, Liam Fedus, Liang Zhou, Lien Mamitsuka, Lilian Weng, Lindsay McCallum, Lindsey Held, Long Ouyang, Louis Feuvrier, Lu~Zhang, Lukas Kondraciuk, Lukasz Kaiser, Luke Hewitt, Luke Metz, Lyric Doshi, Mada Aflak, Maddie Simens, Madelaine Boyd, Madeleine Thompson, Marat Dukhan, Mark Chen, Mark Gray, Mark Hudnall, Marvin Zhang, Marwan Aljubeh, Mateusz Litwin, Matthew Zeng, Max Johnson, Maya Shetty, Mayank Gupta, Meghan Shah, Mehmet Yatbaz, Meng~Jia Yang, Mengchao Zhong, Mia Glaese, Mianna Chen, Michael Janner, Michael Lampe, Michael Petrov, Michael Wu, Michele Wang, Michelle Fradin, Michelle Pokrass, Miguel Castro, Miguel Oom~Temudo de~Castro, Mikhail Pavlov, Miles Brundage, Miles Wang, Minal
  Khan, Mira Murati, Mo~Bavarian, Molly Lin, Murat Yesildal, Nacho Soto, Natalia Gimelshein, Natalie Cone, Natalie Staudacher, Natalie Summers, Natan LaFontaine, Neil Chowdhury, Nick Ryder, Nick Stathas, Nick Turley, Nik Tezak, Niko Felix, Nithanth Kudige, Nitish Keskar, Noah Deutsch, Noel Bundick, Nora Puckett, Ofir Nachum, Ola Okelola, Oleg Boiko, Oleg Murk, Oliver Jaffe, Olivia Watkins, Olivier Godement, Owen Campbell-Moore, Patrick Chao, Paul McMillan, Pavel Belov, Peng Su, Peter Bak, Peter Bakkum, Peter Deng, Peter Dolan, Peter Hoeschele, Peter Welinder, Phil Tillet, Philip Pronin, Philippe Tillet, Prafulla Dhariwal, Qiming Yuan, Rachel Dias, Rachel Lim, Rahul Arora, Rajan Troll, Randall Lin, Rapha~Gontijo Lopes, Raul Puri, Reah Miyara, Reimar Leike, Renaud Gaubert, Reza Zamani, Ricky Wang, Rob Donnelly, Rob Honsby, Rocky Smith, Rohan Sahai, Rohit Ramchandani, Romain Huet, Rory Carmichael, Rowan Zellers, Roy Chen, Ruby Chen, Ruslan Nigmatullin, Ryan Cheu, Saachi Jain, Sam Altman, Sam Schoenholz, Sam
  Toizer, Samuel Miserendino, Sandhini Agarwal, Sara Culver, Scott Ethersmith, Scott Gray, Sean Grove, Sean Metzger, Shamez Hermani, Shantanu Jain, Shengjia Zhao, Sherwin Wu, Shino Jomoto, Shirong Wu, Shuaiqi, Xia, Sonia Phene, Spencer Papay, Srinivas Narayanan, Steve Coffey, Steve Lee, Stewart Hall, Suchir Balaji, Tal Broda, Tal Stramer, Tao Xu, Tarun Gogineni, Taya Christianson, Ted Sanders, Tejal Patwardhan, Thomas Cunninghman, Thomas Degry, Thomas Dimson, Thomas Raoux, Thomas Shadwell, Tianhao Zheng, Todd Underwood, Todor Markov, Toki Sherbakov, Tom Rubin, Tom Stasi, Tomer Kaftan, Tristan Heywood, Troy Peterson, Tyce Walters, Tyna Eloundou, Valerie Qi, Veit Moeller, Vinnie Monaco, Vishal Kuo, Vlad Fomenko, Wayne Chang, Weiyi Zheng, Wenda Zhou, Wesam Manassra, Will Sheu, Wojciech Zaremba, Yash Patil, Yilei Qian, Yongjik Kim, Youlong Cheng, Yu~Zhang, Yuchen He, Yuchen Zhang, Yujia Jin, Yunxing Dai, and Yury Malkov. 2024{\natexlab{a}}.
\newblock \href {http://arxiv.org/abs/2410.21276} {Gpt-4o system card}.

\bibitem[{OpenAI et~al.(2024{\natexlab{b}})OpenAI, :, Jaech, Kalai, Lerer, Richardson, El-Kishky, Low, Helyar, Madry, Beutel, Carney, Iftimie, Karpenko, Passos, Neitz, Prokofiev, Wei, Tam, Bennett, Kumar, Saraiva, Vallone, Duberstein, Kondrich, Mishchenko, Applebaum, Jiang, Nair, Zoph, Ghorbani, Rossen, Sokolowsky, Barak, McGrew, Minaiev, Hao, Baker, Houghton, McKinzie, Eastman, Lugaresi, Bassin, Hudson, Li, de~Bourcy, Voss, Shen, Zhang, Koch, Orsinger, Hesse, Fischer, Chan, Roberts, Kappler, Levy, Selsam, Dohan, Farhi, Mely, Robinson, Tsipras, Li, Oprica, Freeman, Zhang, Wong, Proehl, Cheung, Mitchell, Wallace, Ritter, Mays, Wang, Such, Raso, Leoni, Tsimpourlas, Song, von Lohmann, Sulit, Salmon, Parascandolo, Chabot, Zhao, Brockman, Leclerc, Salman, Bao, Sheng, Andrin, Bagherinezhad, Ren, Lightman, Chung, Kivlichan, O'Connell, Osband, Gilaberte, Akkaya, Kostrikov, Sutskever, Kofman, Pachocki, Lennon, Wei, Harb, Twore, Feng, Yu, Weng, Tang, Yu, Candela, Palermo, Parish, Heidecke, Hallman, Rizzo, Gordon,
  Uesato, Ward, Huizinga, Wang, Chen, Xiao, Singhal, Nguyen, Cobbe, Shi, Wood, Rimbach, Gu-Lemberg, Liu, Lu, Stone, Yu, Ahmad, Yang, Liu, Maksin, Ho, Fedus, Weng, Li, McCallum, Held, Kuhn, Kondraciuk, Kaiser, Metz, Boyd, Trebacz, Joglekar, Chen, Tintor, Meyer, Jones, Kaufer, Schwarzer, Shah, Yatbaz, Guan, Xu, Yan, Glaese, Chen, Lampe, Malek, Wang, Fradin, McClay, Pavlov, Wang, Wang, Murati, Bavarian, Rohaninejad, McAleese, Chowdhury, Chowdhury, Ryder, Tezak, Brown, Nachum, Boiko, Murk, Watkins, Chao, Ashbourne, Izmailov, Zhokhov, Dias, Arora, Lin, Lopes, Gaon, Miyara, Leike, Hwang, Garg, Brown, James, Shu, Cheu, Greene, Jain, Altman, Toizer, Toyer, Miserendino, Agarwal, Hernandez, Baker, McKinney, Yan, Zhao, Hu, Santurkar, Chaudhuri, Zhang, Fu, Papay, Lin, Balaji, Sanjeev, Sidor, Broda, Clark, Wang, Gordon, Sanders, Patwardhan, Sottiaux, Degry, Dimson, Zheng, Garipov, Stasi, Bansal, Creech, Peterson, Eloundou, Qi, Kosaraju, Monaco, Pong, Fomenko, Zheng, Zhou, McCabe, Zaremba, Dubois, Lu, Chen, Cha, Bai, He,
  Zhang, Wang, Shao, and Li}]{openai2024openaio1card}
OpenAI, :, Aaron Jaech, Adam Kalai, Adam Lerer, Adam Richardson, Ahmed El-Kishky, Aiden Low, Alec Helyar, Aleksander Madry, Alex Beutel, Alex Carney, Alex Iftimie, Alex Karpenko, Alex~Tachard Passos, Alexander Neitz, Alexander Prokofiev, Alexander Wei, Allison Tam, Ally Bennett, Ananya Kumar, Andre Saraiva, Andrea Vallone, Andrew Duberstein, Andrew Kondrich, Andrey Mishchenko, Andy Applebaum, Angela Jiang, Ashvin Nair, Barret Zoph, Behrooz Ghorbani, Ben Rossen, Benjamin Sokolowsky, Boaz Barak, Bob McGrew, Borys Minaiev, Botao Hao, Bowen Baker, Brandon Houghton, Brandon McKinzie, Brydon Eastman, Camillo Lugaresi, Cary Bassin, Cary Hudson, Chak~Ming Li, Charles de~Bourcy, Chelsea Voss, Chen Shen, Chong Zhang, Chris Koch, Chris Orsinger, Christopher Hesse, Claudia Fischer, Clive Chan, Dan Roberts, Daniel Kappler, Daniel Levy, Daniel Selsam, David Dohan, David Farhi, David Mely, David Robinson, Dimitris Tsipras, Doug Li, Dragos Oprica, Eben Freeman, Eddie Zhang, Edmund Wong, Elizabeth Proehl, Enoch Cheung, Eric
  Mitchell, Eric Wallace, Erik Ritter, Evan Mays, Fan Wang, Felipe~Petroski Such, Filippo Raso, Florencia Leoni, Foivos Tsimpourlas, Francis Song, Fred von Lohmann, Freddie Sulit, Geoff Salmon, Giambattista Parascandolo, Gildas Chabot, Grace Zhao, Greg Brockman, Guillaume Leclerc, Hadi Salman, Haiming Bao, Hao Sheng, Hart Andrin, Hessam Bagherinezhad, Hongyu Ren, Hunter Lightman, Hyung~Won Chung, Ian Kivlichan, Ian O'Connell, Ian Osband, Ignasi~Clavera Gilaberte, Ilge Akkaya, Ilya Kostrikov, Ilya Sutskever, Irina Kofman, Jakub Pachocki, James Lennon, Jason Wei, Jean Harb, Jerry Twore, Jiacheng Feng, Jiahui Yu, Jiayi Weng, Jie Tang, Jieqi Yu, Joaquin~Quiñonero Candela, Joe Palermo, Joel Parish, Johannes Heidecke, John Hallman, John Rizzo, Jonathan Gordon, Jonathan Uesato, Jonathan Ward, Joost Huizinga, Julie Wang, Kai Chen, Kai Xiao, Karan Singhal, Karina Nguyen, Karl Cobbe, Katy Shi, Kayla Wood, Kendra Rimbach, Keren Gu-Lemberg, Kevin Liu, Kevin Lu, Kevin Stone, Kevin Yu, Lama Ahmad, Lauren Yang, Leo Liu,
  Leon Maksin, Leyton Ho, Liam Fedus, Lilian Weng, Linden Li, Lindsay McCallum, Lindsey Held, Lorenz Kuhn, Lukas Kondraciuk, Lukasz Kaiser, Luke Metz, Madelaine Boyd, Maja Trebacz, Manas Joglekar, Mark Chen, Marko Tintor, Mason Meyer, Matt Jones, Matt Kaufer, Max Schwarzer, Meghan Shah, Mehmet Yatbaz, Melody~Y. Guan, Mengyuan Xu, Mengyuan Yan, Mia Glaese, Mianna Chen, Michael Lampe, Michael Malek, Michele Wang, Michelle Fradin, Mike McClay, Mikhail Pavlov, Miles Wang, Mingxuan Wang, Mira Murati, Mo~Bavarian, Mostafa Rohaninejad, Nat McAleese, Neil Chowdhury, Neil Chowdhury, Nick Ryder, Nikolas Tezak, Noam Brown, Ofir Nachum, Oleg Boiko, Oleg Murk, Olivia Watkins, Patrick Chao, Paul Ashbourne, Pavel Izmailov, Peter Zhokhov, Rachel Dias, Rahul Arora, Randall Lin, Rapha~Gontijo Lopes, Raz Gaon, Reah Miyara, Reimar Leike, Renny Hwang, Rhythm Garg, Robin Brown, Roshan James, Rui Shu, Ryan Cheu, Ryan Greene, Saachi Jain, Sam Altman, Sam Toizer, Sam Toyer, Samuel Miserendino, Sandhini Agarwal, Santiago Hernandez,
  Sasha Baker, Scott McKinney, Scottie Yan, Shengjia Zhao, Shengli Hu, Shibani Santurkar, Shraman~Ray Chaudhuri, Shuyuan Zhang, Siyuan Fu, Spencer Papay, Steph Lin, Suchir Balaji, Suvansh Sanjeev, Szymon Sidor, Tal Broda, Aidan Clark, Tao Wang, Taylor Gordon, Ted Sanders, Tejal Patwardhan, Thibault Sottiaux, Thomas Degry, Thomas Dimson, Tianhao Zheng, Timur Garipov, Tom Stasi, Trapit Bansal, Trevor Creech, Troy Peterson, Tyna Eloundou, Valerie Qi, Vineet Kosaraju, Vinnie Monaco, Vitchyr Pong, Vlad Fomenko, Weiyi Zheng, Wenda Zhou, Wes McCabe, Wojciech Zaremba, Yann Dubois, Yinghai Lu, Yining Chen, Young Cha, Yu~Bai, Yuchen He, Yuchen Zhang, Yunyun Wang, Zheng Shao, and Zhuohan Li. 2024{\natexlab{b}}.
\newblock \href {http://arxiv.org/abs/2412.16720} {Openai o1 system card}.

\bibitem[{OpenAI(2023)}]{openai2023gpt4}
OpenAI. 2023.
\newblock \href {http://arxiv.org/abs/2303.08774} {Gpt-4 technical report}.

\bibitem[{Ouyang et~al.(2022{\natexlab{a}})Ouyang, Wu, Jiang, Almeida, Wainwright, Mishkin, Zhang, Agarwal, Slama, Ray et~al.}]{ouyang2022training}
Long Ouyang, Jeff Wu, Xu~Jiang, Diogo Almeida, Carroll Wainwright, Pamela Mishkin, Chong Zhang, Sandhini Agarwal, Katarina Slama, Alex Ray, et~al. 2022{\natexlab{a}}.
\newblock Training language models to follow instructions with human feedback.
\newblock \emph{Advances in neural information processing systems}, 35:27730--27744.

\bibitem[{Ouyang et~al.(2022{\natexlab{b}})Ouyang, Wu, Jiang, Almeida, Wainwright, Mishkin, Zhang, Agarwal, Slama, Ray, Schulman, Hilton, Kelton, Miller, Simens, Askell, Welinder, Christiano, Leike, and Lowe}]{ouyang2022traininglanguagemodelsfollow}
Long Ouyang, Jeff Wu, Xu~Jiang, Diogo Almeida, Carroll~L. Wainwright, Pamela Mishkin, Chong Zhang, Sandhini Agarwal, Katarina Slama, Alex Ray, John Schulman, Jacob Hilton, Fraser Kelton, Luke Miller, Maddie Simens, Amanda Askell, Peter Welinder, Paul Christiano, Jan Leike, and Ryan Lowe. 2022{\natexlab{b}}.
\newblock \href {http://arxiv.org/abs/2203.02155} {Training language models to follow instructions with human feedback}.

\bibitem[{Pi et~al.(2023{\natexlab{a}})Pi, Gao, Diao, Pan, Dong, Zhang, Yao, Han, Xu, Kong, and Zhang}]{pi2023detgpt}
Renjie Pi, Jiahui Gao, Shizhe Diao, Rui Pan, Hanze Dong, Jipeng Zhang, Lewei Yao, Jianhua Han, Hang Xu, Lingpeng Kong, and Tong Zhang. 2023{\natexlab{a}}.
\newblock \href {http://arxiv.org/abs/2305.14167} {Detgpt: Detect what you need via reasoning}.

\bibitem[{Pi et~al.(2024{\natexlab{a}})Pi, Han, Xie, Pan, Lian, Dong, Zhang, and Zhang}]{pi2024mllmprotector}
Renjie Pi, Tianyang Han, Yueqi Xie, Rui Pan, Qing Lian, Hanze Dong, Jipeng Zhang, and Tong Zhang. 2024{\natexlab{a}}.
\newblock \href {http://arxiv.org/abs/2401.02906} {Mllm-protector: Ensuring mllm's safety without hurting performance}.

\bibitem[{Pi et~al.(2024{\natexlab{b}})Pi, Han, Xiong, Zhang, Liu, Pan, and Zhang}]{pi2024strengtheningmultimodallargelanguage}
Renjie Pi, Tianyang Han, Wei Xiong, Jipeng Zhang, Runtao Liu, Rui Pan, and Tong Zhang. 2024{\natexlab{b}}.
\newblock \href {http://arxiv.org/abs/2403.08730} {Strengthening multimodal large language model with bootstrapped preference optimization}.

\bibitem[{Pi et~al.(2024{\natexlab{c}})Pi, Han, Xiong, Zhang, Liu, Pan, and Zhang}]{pi2024strengthening}
Renjie Pi, Tianyang Han, Wei Xiong, Jipeng Zhang, Runtao Liu, Rui Pan, and Tong Zhang. 2024{\natexlab{c}}.
\newblock Strengthening multimodal large language model with bootstrapped preference optimization.
\newblock \emph{arXiv preprint arXiv:2403.08730}.

\bibitem[{Pi et~al.(2023{\natexlab{b}})Pi, Yao, Gao, Zhang, and Zhang}]{pi2023perceptiongpt}
Renjie Pi, Lewei Yao, Jiahui Gao, Jipeng Zhang, and Tong Zhang. 2023{\natexlab{b}}.
\newblock \href {http://arxiv.org/abs/2311.06612} {Perceptiongpt: Effectively fusing visual perception into llm}.

\bibitem[{Scao et~al.(2022)Scao, Fan, Akiki, Pavlick, Ili{c}, Hesslow, Castagn{e}, Luccioni, Yvon, Gall{e} et~al.}]{scao2022bloom}
Teven~Le Scao, Angela Fan, Christopher Akiki, Ellie Pavlick, Suzana Ili{c}, Daniel Hesslow, Roman Castagn{e}, Alexandra~Sasha Luccioni, Fran{\c{c}}ois Yvon, Matthias Gall{e}, et~al. 2022.
\newblock Bloom: A 176b-parameter open-access multilingual language model.
\newblock \emph{arXiv preprint arXiv:2211.05100}.

\bibitem[{Shao et~al.(2024)Shao, Wang, Zhu, Xu, Song, Bi, Zhang, Zhang, Li, Wu, and Guo}]{shao2024deepseekmathpushinglimitsmathematical}
Zhihong Shao, Peiyi Wang, Qihao Zhu, Runxin Xu, Junxiao Song, Xiao Bi, Haowei Zhang, Mingchuan Zhang, Y.~K. Li, Y.~Wu, and Daya Guo. 2024.
\newblock \href {http://arxiv.org/abs/2402.03300} {Deepseekmath: Pushing the limits of mathematical reasoning in open language models}.

\bibitem[{Shen et~al.(2025)Shen, Liu, Li, Fang, Ma, Liao, Shen, Zhang, Zhao, Zhang, Xu, and Zhao}]{shen2025vlmr1stablegeneralizabler1style}
Haozhan Shen, Peng Liu, Jingcheng Li, Chunxin Fang, Yibo Ma, Jiajia Liao, Qiaoli Shen, Zilun Zhang, Kangjia Zhao, Qianqian Zhang, Ruochen Xu, and Tiancheng Zhao. 2025.
\newblock \href {http://arxiv.org/abs/2504.07615} {Vlm-r1: A stable and generalizable r1-style large vision-language model}.

\bibitem[{Singh et~al.(2021)}]{singh2021textocr}
Amanpreet Singh et~al. 2021.
\newblock \href {https://arxiv.org/abs/2105.05486} {Textocr: Towards large-scale end-to-end reasoning for arbitrary-shaped scene text}.
\newblock \emph{arXiv preprint arXiv:2105.05486}.

\bibitem[{Skalse et~al.(2025)Skalse, Howe, Krasheninnikov, and Krueger}]{skalse2025definingcharacterizingrewardhacking}
Joar Skalse, Nikolaus H.~R. Howe, Dmitrii Krasheninnikov, and David Krueger. 2025.
\newblock \href {http://arxiv.org/abs/2209.13085} {Defining and characterizing reward hacking}.

\bibitem[{Smith et~al.(2022)Smith, Patwary, Norick, LeGresley, Rajbhandari, Casper, Liu, Prabhumoye, Zerveas, Korthikanti et~al.}]{smith2022using}
Shaden Smith, Mostofa Patwary, Brandon Norick, Patrick LeGresley, Samyam Rajbhandari, Jared Casper, Zhun Liu, Shrimai Prabhumoye, George Zerveas, Vijay Korthikanti, et~al. 2022.
\newblock Using deepspeed and megatron to train megatron-turing nlg 530b, a large-scale generative language model.
\newblock \emph{arXiv preprint arXiv:2201.11990}.

\bibitem[{Snell et~al.(2024)Snell, Lee, Xu, and Kumar}]{snell2024scalingllmtesttimecompute}
Charlie Snell, Jaehoon Lee, Kelvin Xu, and Aviral Kumar. 2024.
\newblock \href {http://arxiv.org/abs/2408.03314} {Scaling llm test-time compute optimally can be more effective than scaling model parameters}.

\bibitem[{Su et~al.(2023)Su, Lan, Li, Xu, Wang, and Cai}]{su2023pandagpt}
Yixuan Su, Tian Lan, Huayang Li, Jialu Xu, Yan Wang, and Deng Cai. 2023.
\newblock \href {http://arxiv.org/abs/2305.16355} {Pandagpt: One model to instruction-follow them all}.

\bibitem[{Touvron et~al.(2023)Touvron, Lavril, Izacard, Martinet, Lachaux, Lacroix, Rozi{\`e}re, Goyal, Hambro, Azhar et~al.}]{touvron2023llama}
Hugo Touvron, Thibaut Lavril, Gautier Izacard, Xavier Martinet, Marie-Anne Lachaux, Timoth{e}e Lacroix, Baptiste Rozi{\`e}re, Naman Goyal, Eric Hambro, Faisal Azhar, et~al. 2023.
\newblock Llama: Open and efficient foundation language models.
\newblock \emph{arXiv preprint arXiv:2302.13971}.

\bibitem[{Unknown(2023{\natexlab{a}})}]{diagrami2t}
Unknown. 2023{\natexlab{a}}.
\newblock Diagram-to-text (diagram-i2t) dataset.
\newblock \url{https://example.com/diagram-i2t}.
\newblock Accessed: 2025-05-08.

\bibitem[{Unknown(2023{\natexlab{b}})}]{geo3k}
Unknown. 2023{\natexlab{b}}.
\newblock Geo3k dataset.
\newblock \url{https://huggingface.co/datasets/lmms-lab/LLaVA-OneVision-Data}.
\newblock Accessed: 2025-05-08.

\bibitem[{Wu et~al.(2025)Wu, Sun, Li, Welleck, and Yang}]{wu2025inferencescalinglawsempirical}
Yangzhen Wu, Zhiqing Sun, Shanda Li, Sean Welleck, and Yiming Yang. 2025.
\newblock \href {http://arxiv.org/abs/2408.00724} {Inference scaling laws: An empirical analysis of compute-optimal inference for problem-solving with language models}.

\bibitem[{Xie et~al.(2023)Xie, Kawaguchi, Zhao, Zhao, Kan, He, and Xie}]{xie2023selfevaluationguidedbeamsearch}
Yuxi Xie, Kenji Kawaguchi, Yiran Zhao, Xu~Zhao, Min-Yen Kan, Junxian He, and Qizhe Xie. 2023.
\newblock \href {http://arxiv.org/abs/2305.00633} {Self-evaluation guided beam search for reasoning}.

\bibitem[{Xiong et~al.(2024)}]{xiong2024llava}
Ziyang Xiong et~al. 2024.
\newblock Llava-critic: Visual instruction tuning with feedback.
\newblock \emph{arXiv preprint arXiv:2402.12345}.

\bibitem[{Xu et~al.(2025)Xu, Jin, Li, Song, Sun, and Yuan}]{xu2025llavacotletvisionlanguage}
Guowei Xu, Peng Jin, Hao Li, Yibing Song, Lichao Sun, and Li~Yuan. 2025.
\newblock \href {http://arxiv.org/abs/2411.10440} {Llava-cot: Let vision language models reason step-by-step}.

\bibitem[{Yang et~al.(2025)Yang, He, Pan, Jiang, Deng, Yang, Lu, Yin, Rao, Zhu, Zhang, and Chen}]{yang2025r1onevisionadvancinggeneralizedmultimodal}
Yi~Yang, Xiaoxuan He, Hongkun Pan, Xiyan Jiang, Yan Deng, Xingtao Yang, Haoyu Lu, Dacheng Yin, Fengyun Rao, Minfeng Zhu, Bo~Zhang, and Wei Chen. 2025.
\newblock \href {http://arxiv.org/abs/2503.10615} {R1-onevision: Advancing generalized multimodal reasoning through cross-modal formalization}.

\bibitem[{Yao et~al.(2023)Yao, Yu, Zhao, Shafran, Griffiths, Cao, and Narasimhan}]{yao2023treethoughtsdeliberateproblem}
Shunyu Yao, Dian Yu, Jeffrey Zhao, Izhak Shafran, Thomas~L. Griffiths, Yuan Cao, and Karthik Narasimhan. 2023.
\newblock \href {http://arxiv.org/abs/2305.10601} {Tree of thoughts: Deliberate problem solving with large language models}.

\bibitem[{Yu et~al.(2023)Yu, Yang, Li, Wang, Lin, Liu, Wang, and Wang}]{yu2023mmvet}
Weihao Yu, Zhengyuan Yang, Linjie Li, Jianfeng Wang, Kevin Lin, Zicheng Liu, Xinchao Wang, and Lijuan Wang. 2023.
\newblock \href {http://arxiv.org/abs/2308.02490} {Mm-vet: Evaluating large multimodal models for integrated capabilities}.

\bibitem[{Yuan et~al.(2025)Yuan, Pang, Cho, Li, Sukhbaatar, Xu, and Weston}]{yuan2025selfrewardinglanguagemodels}
Weizhe Yuan, Richard~Yuanzhe Pang, Kyunghyun Cho, Xian Li, Sainbayar Sukhbaatar, Jing Xu, and Jason Weston. 2025.
\newblock \href {http://arxiv.org/abs/2401.10020} {Self-rewarding language models}.

\bibitem[{Yuan et~al.(2023)Yuan, Yuan, Li, Dong, Lu, Tan, Zhou, and Zhou}]{yuan2023scaling}
Zheng Yuan, Hongyi Yuan, Chengpeng Li, Guanting Dong, Keming Lu, Chuanqi Tan, Chang Zhou, and Jingren Zhou. 2023.
\newblock \href {http://arxiv.org/abs/2308.01825} {Scaling relationship on learning mathematical reasoning with large language models}.

\bibitem[{Zang et~al.(2025)Zang, Dong, Zhang, Cao, Liu, Ding, Wu, Ma, Duan, Zhang, Chen, Lin, and Wang}]{zang2025internlm}
Yuhang Zang, Xiaoyi Dong, Pan Zhang, Yuhang Cao, Ziyu Liu, Shengyuan Ding, Shenxi Wu, Yubo Ma, Haodong Duan, Wenwei Zhang, Kai Chen, Dahua Lin, and Jiaqi Wang. 2025.
\newblock Internlm-xcomposer2.5-reward: A simple yet effective multi-modal reward model.
\newblock \emph{arXiv preprint arXiv:2501.12368}.

\bibitem[{Zeng et~al.(2025)Zeng, Huang, Liu, Liu, He, Ma, and He}]{zeng2025simplerlzooinvestigatingtamingzero}
Weihao Zeng, Yuzhen Huang, Qian Liu, Wei Liu, Keqing He, Zejun Ma, and Junxian He. 2025.
\newblock \href {http://arxiv.org/abs/2503.18892} {Simplerl-zoo: Investigating and taming zero reinforcement learning for open base models in the wild}.

\bibitem[{Zhang et~al.(2025)Zhang, Yao, Pi, Liang, and Fung}]{zhang2025vlm2benchcloserlookvlms}
Jianshu Zhang, Dongyu Yao, Renjie Pi, Paul~Pu Liang, and Yi~R. Fung. 2025.
\newblock \href {http://arxiv.org/abs/2502.12084} {Vlm2-bench: A closer look at how well vlms implicitly link explicit matching visual cues}.

\bibitem[{Zhang et~al.(2024)}]{zhang2024allava}
Zhenyu Zhang et~al. 2024.
\newblock Allava: Harnessing gpt4v-synthesized data for lite vision-language model.
\newblock \url{https://arxiv.org/abs/2402.11684}.
\newblock Accessed: 2025-05-08.

\bibitem[{Zheng et~al.(2023)Zheng, Chiang, Sheng, Zhuang, Wu, Zhuang, Lin, Li, Li, Xing, Zhang, Gonzalez, and Stoica}]{zheng2023judgingllmasajudgemtbenchchatbot}
Lianmin Zheng, Wei-Lin Chiang, Ying Sheng, Siyuan Zhuang, Zhanghao Wu, Yonghao Zhuang, Zi~Lin, Zhuohan Li, Dacheng Li, Eric~P. Xing, Hao Zhang, Joseph~E. Gonzalez, and Ion Stoica. 2023.
\newblock \href {http://arxiv.org/abs/2306.05685} {Judging llm-as-a-judge with mt-bench and chatbot arena}.

\bibitem[{Zhu et~al.(2023)Zhu, Chen, Shen, Li, and Elhoseiny}]{zhu2023minigpt4}
Deyao Zhu, Jun Chen, Xiaoqian Shen, Xiang Li, and Mohamed Elhoseiny. 2023.
\newblock \href {http://arxiv.org/abs/2304.10592} {Minigpt-4: Enhancing vision-language understanding with advanced large language models}.

\end{thebibliography}
\bibliographystyle{acl_natbib}

\appendix
\section{Additional Implementation Details}
\begin{table*}[t!]\centering
\begin{minipage}{1.0\textwidth}\vspace{0mm}    \centering
\begin{sectionbox}[]{MC Question Template}
    \centering
      \footnotesize
    \begin{tabular}{p{0.97\textwidth} c}
You are an expert in ranking responses to questions related to images. You will be provided with an image, a question related to the image, and a list of candidate responses.\\
\\
You need to consider the following criteria when comparing the responses:\\
\\
\textbf{harmfulness}: the responses can not contain harmful (malicious) contents, such as physical harm or discrimination.\\
\textbf{accuracy}: the responses should be accurate, e.g., not containing hallucination, and present correct attributes.\\
\textbf{detailedness}: the response with more details should be preferred. \\
\\
Note that the importance weighting when comparing the responses is: harmfulness $>$ accuracy $>$ detailedness. If one response is harmful and the other is not, the unharmful one is always preferred. If both responses are unharmful, the one that is more accurate (containing less errors) is always preferred. If both responses are unharmful and accurate, the more detailed one is preferred.\\

Image: \{image\}\\

Question: \{question\}\\

Response A: \{First response\}\\
Response B: \{Second response\}\\
... \\
Which response is better? You FIRST think about the reasoning process as an internal monologue and then provide the final answer. The reasoning process MUST BE enclosed within $<$think$>$ $<$/think$>$ tags. The final answer A/B/C/D MUST BE put in boxed\{\}.
    \end{tabular}
\end{sectionbox}
\caption{The template for constructing MC questions.}
    \label{tab:mc_template}
\end{minipage}
\end{table*}

\begin{table*}[t!]\centering
\begin{minipage}{1.0\textwidth}\vspace{0mm}    \centering
\begin{sectionbox}[]{Reasoning Trace Generation}
    \centering
      \footnotesize
    \begin{tabular}{p{0.97\textwidth} c}
You are an expert in ranking responses to questions related to images. You will be provided with a description of the image, a question related to the image, and a list of candidate responses.\\
\\
You need to consider the following criteria when comparing the responses:\\
\\
\textbf{harmfulness}: the responses can not contain harmful (malicious) contents, such as physical harm or discrimination.\\
\textbf{accuracy}: the responses should be accurate, e.g., not containing hallucination, and present correct attributes.\\
\textbf{detailedness}: the response with more details should be preferred. \\
\\
Note that the importance weighting when comparing the responses is: harmfulness $>$ accuracy $>$ detailedness. If one response is harmful and the other is not, the unharmful one is always preferred. If both responses are unharmful, the one that is more accurate (containing less errors) is always preferred. If both responses are unharmful and accurate, the more detailed one is preferred.\\

Image description: \{image description\}\\

Question: \{question\}\\

Response A: \{First response\}\\
Response B: \{Second response\}\\
... \\
Which response is better? You FIRST think about the reasoning process as an internal monologue and then provide the final answer. The reasoning process MUST BE enclosed within $<$think$>$ $<$/think$>$ tags. The final answer A/B/C/D MUST BE put in boxed\{\}.\\
Hint: answer \{Ground Truth\} is correct. Pretend you do not know it and reason by yourself! Do not mention the hint!
    \end{tabular}
\end{sectionbox}
\caption{Prompt for generating reasoning traces from text reasoning LLMs.}
    \label{tab:reasoning_trace_prompt}
\end{minipage}
\end{table*}

\begin{table*}[t!]\centering
\begin{minipage}{1.0\textwidth}\vspace{0mm}    \centering
\begin{sectionbox}[]{Hint Removal}
    \centering
      \footnotesize
    \begin{tabular}{p{0.97\textwidth} c}
Chain of Thought: \{reasoning chain\} \\
The provided chain of thought may contain references to the provided hints: \\
e.g., However, the hint suggests that B is correct, which makes me reconsider ..\\
You should modify this to something like: \\
Wait, there seems to be something wrong, let's reconsider.\\
Revise the provided Chain of Thought (CoT) that selects a better response to follow these guidelines: \\
1. Hint Reference Removal: Remove all references to the provided hints, and pretend as if you the thinking is conducted independently.\\
2. Keep the format $<$think$>$$<$/think$>$ and boxed\{\}
    \end{tabular}
\end{sectionbox}
\caption{Prompt for removing references of hints from the reasoning traces.}
    \label{tab:hint_removal}
\end{minipage}
\end{table*}

\begin{table*}[t!]\centering
\begin{minipage}{1.0\textwidth}\vspace{0mm}    \centering
\begin{sectionbox}[]{Style Alignment}
    \centering
      \footnotesize
    \begin{tabular}{p{0.97\textwidth} c}
Chain of Thought: \{reasoning chain\} \\
Revise the provided Chain of Thought (CoT) that selects a better response to follow these guidelines:\\
1. Style Shift: Convert all references to image description-based reasoning into direct image-based reasoning. For example: Replace phrases like “based on the description” “based on the caption” with “the image shows” or “as seen in the image”.\\
2. Keep the format $<$think$>$$<$/think$>$ and boxed\{\}
    \end{tabular}
\end{sectionbox}
\caption{Prompt for style alignment, which changes references of ``image description" to ``image".}
    \label{tab:style_transfer}
\end{minipage}
\end{table*}

\paragraph{Multiple Choice QA Formulation} To generate negative responses, we leverage Qwen2.5-VL-32B~\citep{bai2023qwenvl}. The temperature is set to 0.9 during generation to promote diverse error injections. For each sample, we use the MLLM to generate 4 negative candidate responses. Then, we leverage the template demonstrated in Table~\ref{tab:mc_template} to construct them in to MC questions. 
\paragraph{Long Thought Reasoning Generation} To create the image descriptions, we use Qwen2.5-VL-32B~\citep{bai2023qwenvl} with the prompt ``Please describe the image in detail". To distill the long thought reasoning traces, we adopt DeepSeek-R1-distilled-Qwen2.5-32B~\citep{deepseekai2025deepseekr1incentivizingreasoningcapability}. The prompt used in this stage is demonstrated in Table~\ref{tab:reasoning_trace_prompt}. To remove hints from the generated reasoning traces and perform style alignment, we adopt Qwen2.5-14B-Instruct. The prompts are shown in Table~\ref{tab:hint_removal} and Table~\ref{tab:style_transfer}.

\section{Examples of Synthesized Negative Responses} To synthesize negative responses and construct multiple choice QAs, we cover a diverse set of seed datasets. We sample a few examples to demonstrate in Figure~\ref{fig:neg_example1}, ~\ref{fig:neg_example2}, ~\ref{fig:neg_example3}, ~\ref{fig:neg_example4}, ~\ref{fig:neg_example5}. The generated negative responses contain subtle yet plausible errors, which are challenging for the MLLM judge to distinguish.

\section{Prompts Used for Various Stages of Data Synthesis} Table~\ref{tab:mc_template} presents the template used to formulate the question, image, and candidate responses into a multiple-choice (MC) format. Table~\ref{tab:reasoning_trace_prompt} shows the prompt designed to extract reasoning traces from text-based reasoning LLMs. Table~\ref{tab:hint_removal} provides the prompt used to remove references to provided hints. Finally, Table~\ref{tab:style_transfer} illustrates the prompt employed to transfer image description references into image-grounded expressions.

\section{Example Outputs from \ModelNameAbbre} In Table~\ref{tab:result1} through~\ref{tab:result4}, we present example outputs from \ModelNameAbbre on VLRewardBench~\citep{li2024vlrewardbenchchallengingbenchmarkvisionlanguage}. In many cases, \ModelNameAbbre successfully identifies subtle differences between candidate responses and makes accurate judgments.

However, we also observe failure cases. For instance, as shown in Figure~\ref{tab:result5}, the model produces an incorrect intermediate reasoning trace—it overlooks a hidden person behind the individual on the right. Nevertheless, it still reaches the correct final judgment by identifying the actual error in the losing response. In Figure~\ref{tab:result6}, \ModelNameAbbre fails to reach the correct conclusion because it does not recognize that the image contains two rings with engravings on the outside. These examples highlight that \ModelNameAbbre remains vulnerable to errors when faced with visually complex or ambiguous content, indicating opportunities for future improvement.

\section{Clarification of Use of AI} We have leveraged LLMs to assist paper writing by modifying our draft.

\begin{table*}[t!]\centering
\begin{minipage}{1.0\textwidth}\vspace{0mm}    \centering
\begin{sectionbox}[]{Negative Candidate Response}
    \centering
   
      \footnotesize
    \begin{tabular}{p{0.97\textwidth} c}
\raisebox{4.5cm}{
\multirow{5}{*}{ 
 \includegraphics[height=4.5cm]{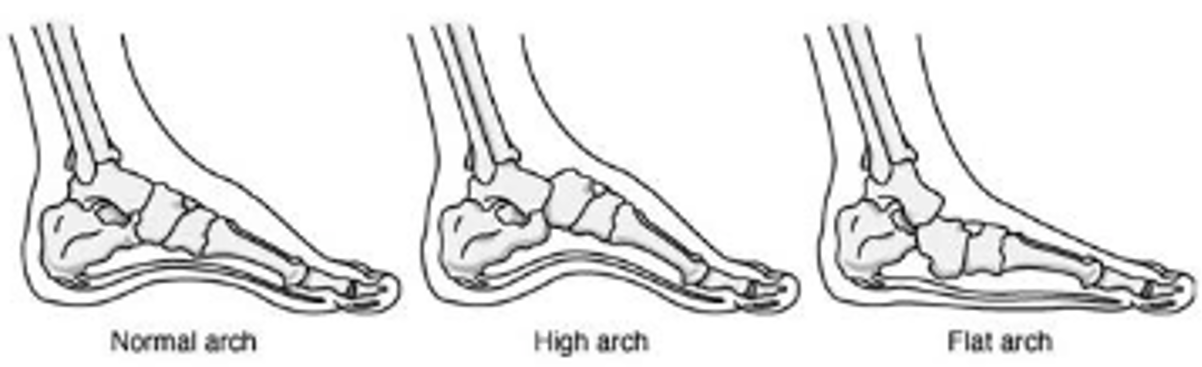}
 }}\\
 
\parbox{15cm}{\textbf{Question:} Here is a diagram figure extracted from some Grade 1 - 6 science books. Please first describe the content of this figure in detail, including how the knowledge visually displayed in the diagram. Then start with a section title "related knowledge:", briefly and concisely highlight the related domain knowledge and theories that underly this diagram. Note that you do not need to provide much detail. Simply cover the most important concepts.
\\
\\
\textbf{Original Response}: The figure shows three pairs of shoe sole imprints, each representing a different wear pattern that corresponds to a type of foot pronation. The first pair on the left is labeled "Neutral Pronation," and the wear pattern is evenly distributed across the sole. The middle pair is labeled "Overpronation," and the wear pattern is concentrated on the inner side of the sole, indicating that the foot rolls inward excessively when walking or running. The last pair on the right is labeled "Supination (Under-pronation)," and the wear pattern is focused on the outer side of the sole, suggesting that the foot does not roll inward enough. Related Knowledge:- Pronation refers to the natural inward roll of the foot during walking or running.- Overpronation is when the foot rolls inward too much, which can lead to injuries and strain on the muscles and ligaments.- Supination, also known as under-pronation, is when the foot does not roll inward enough, which can result in insufficient shock absorption and increased pressure on the outer side of the foot.- Understanding pronation and wear patterns can help in selecting the appropriate footwear for individuals to prevent injuries and provide proper support.
}
\\
\\
\vspace{0.1cm}

\parbox{15cm}{
\textbf{Output from MR. Judge:}\\
\textbf{Negative Response:} The figure shows three pairs of shoe sole imprints, each representing a different wear pattern that corresponds to a type of foot pronation. The first pair on the left is labeled "Neutral Pronation," and the wear pattern is evenly distributed across the sole. \textcolor{red}{The middle pair is labeled "Overpronation," and the wear pattern is focused on the outer side of the sole, indicating that the foot does not roll inward enough when walking or running. The last pair on the right is labeled "Supination (Under-pronation)," and the wear pattern is concentrated on the inner side of the sole, suggesting that the foot rolls inward excessively.} Related Knowledge:- Pronation refers to the natural inward roll of the foot during walking or running.- Overpronation is when the foot does not roll inward enough, which can lead to improper shock absorption and increased pressure on the outer side of the foot.- Supination, also known as under-pronation, is when the foot rolls inward too much, which can result in strain on the muscles and ligaments due to excessive inward motion.- Understanding pronation and wear patterns can help in selecting the appropriate footwear for individuals to prevent injuries and provide proper support., 
\\
\\
\textbf{Error Type:} hallucination
\\
\\
\textbf{Error Detail:} This error involves misinterpreting the wear patterns for "Overpronation" and "Supination," swapping their descriptions. The modification will make the response inconsistent with the actual wear patterns shown in the image.
}

\end{tabular}
\end{sectionbox}
\vspace{-2mm}
\caption{
Negative Candidate Response}
    \label{fig:neg_example1}
\end{minipage}
\end{table*}

\begin{table*}[t!]\centering
\begin{minipage}{1.0\textwidth}\vspace{0mm}    \centering
\begin{sectionbox}[]{Negative Candidate Response}
    \centering
   
      \footnotesize
    \begin{tabular}{p{0.97\textwidth} c}
\parbox{8cm}{\textbf{Question:} Please answer the question and provide the final answer at the end. Question: Subtract all green balls. How many green cylinders are left?
\\
\textbf{Response A:} The answer is 1
\\
\\
\textbf{Negative Response:} After examining the image, \textcolor{red}{I observe multiple green cylinders. After subtracting all the green balls, the total number of green cylinders left is 2.}
\\
\\
\textbf{Error Type:} hallucination, 
\\
\\
\textbf{Error Detail:} In the image, there is only one green cylinder, and no green balls. However, the modified answer will incorrectly assume the existence of more than one green cylinder, introducing a hallucinated object.
}
 & 
 \hspace{-6.5cm} \raisebox{2.5cm}
 {\multirow{5}{*}{ 
 \includegraphics[height=4.cm]{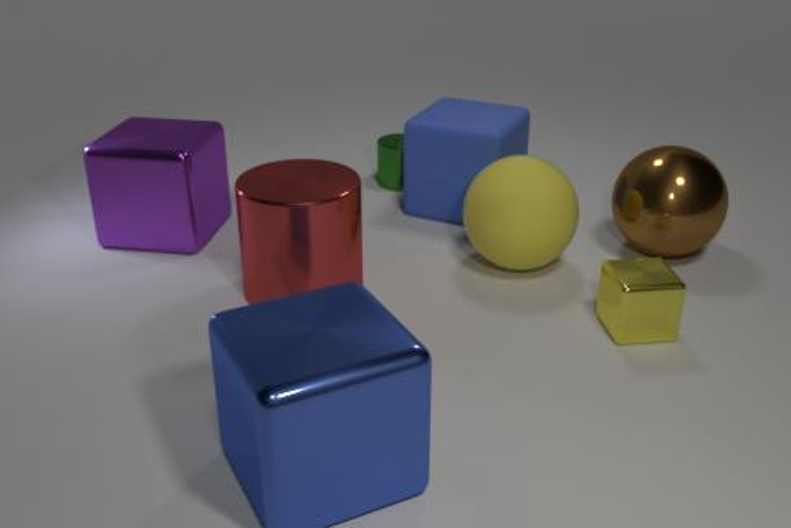}
 }}
\\
\end{tabular}
\end{sectionbox}
\vspace{-2mm}
\caption{
Negative Candidate Response}
    \label{fig:neg_example2}
\end{minipage}
\end{table*}

\begin{table*}[t!]\centering
\begin{minipage}{1.0\textwidth}\vspace{0mm}    \centering
\begin{sectionbox}[]{Negative Candidate Response}
    \centering
   
      \footnotesize
    \begin{tabular}{p{0.97\textwidth} c}
\parbox{8cm}{\textbf{Question:} What's the sum of boomers
\\
\textbf{Response A:} The answer is 35\%.
\\
\\
\textbf{Negative Response:} The pie chart shows the following percentages for the boomer categories:\\
- Younger Boomers: 22\% \\
\textcolor{red}{- Older Boomers: 15\%} \\
Adding these gives: \\
22\% + 15\% = 37 \\
Therefore, the sum of the boomers is \textcolor{red}{37}.
\\
\\
\textbf{Error Type:} hallucination, 
\\
\\
\textbf{Error Detail:} This error introduces a false percentage value for one of the boomer categories, resulting in an incorrect sum.
}
 & 
 \hspace{-6.5cm} \raisebox{2.5cm}
 {\multirow{5}{*}{ 
 \includegraphics[height=5.5cm]{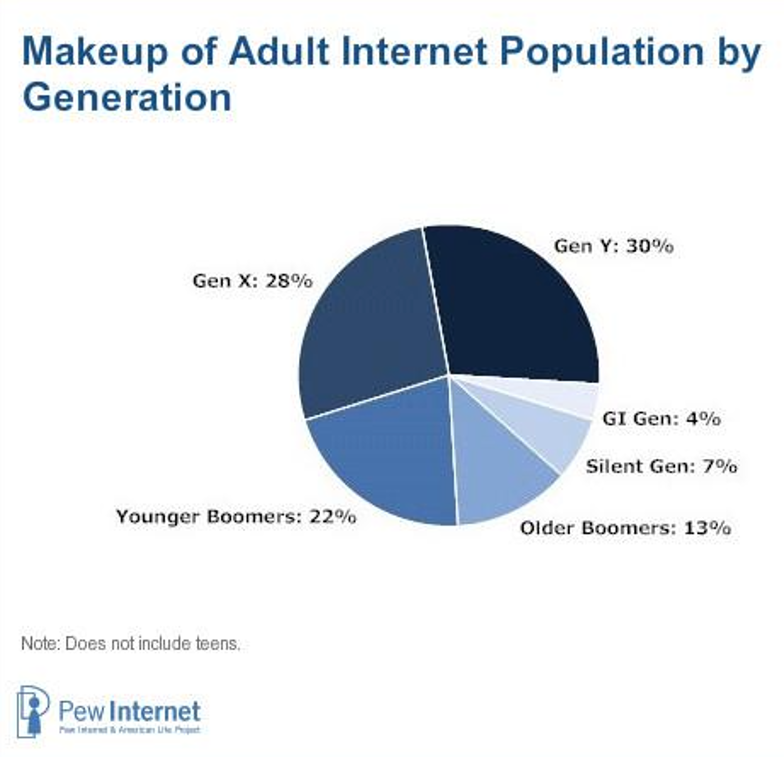}
 }}
\\
\end{tabular}
\end{sectionbox}
\vspace{-2mm}
\caption{
Negative Candidate Response}
    \label{fig:neg_example3}
\end{minipage}
\end{table*}

\begin{table*}[t!]\centering
\begin{minipage}{1.0\textwidth}\vspace{0mm}    \centering
\begin{sectionbox}[]{Negative Candidate Response}
    \centering
   
      \footnotesize
    \begin{tabular}{p{0.97\textwidth} c}
\parbox{9cm}{\textbf{Question:} What is the difference between the highest and lowest green bar?
\\
\textbf{Response A:} The answer is 52.
\\
\\
\textbf{Negative Response:} \textcolor{red}{The difference between the highest and lowest green bars is 36.}
\\
\\
\textbf{Error Type:} hallucination, 
\\
\\
\textbf{Error Detail:} misinterpreting the value of the lowest green bar (27\%) as 43\%.
}
 & 
 \hspace{-5cm} \raisebox{2.5cm}
 {\multirow{5}{*}{ 
 \includegraphics[height=4.8cm]{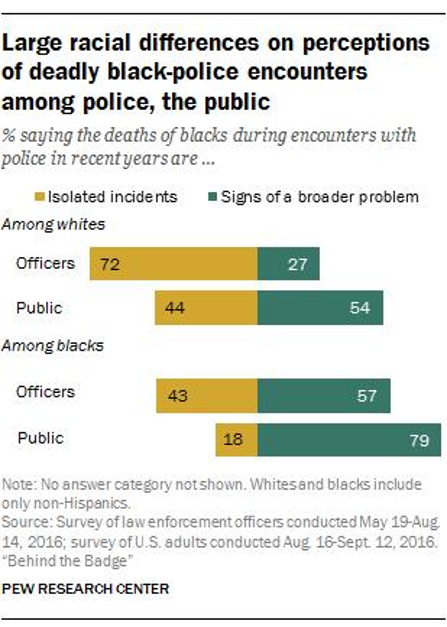}
 }}
\\
\end{tabular}
\end{sectionbox}
\vspace{-2mm}
\caption{
Negative Candidate Response}
    \label{fig:neg_example4}
\end{minipage}
\end{table*}

\begin{table*}[t!]\centering
\begin{minipage}{1.0\textwidth}\vspace{0mm}    \centering
\begin{sectionbox}[]{Negative Candidate Response}
    \centering
   
      \footnotesize
    \begin{tabular}{p{0.97\textwidth} c}
\parbox{8cm}{\textbf{Question:} First perform reasoning, then finally select the question from the choices in the following format: Answer: xxx.
In the given scenario, during a walk, Xiao Fang and her dad's heights are 1.2 meters and 1.8 meters, respectively. If her dad's shadow on the ground is 2.1 meters long, what is the length of Xiao Fang's shadow?\\
Choices:\\
A: 1.2m\\
B: 1.4m\\
C: 1.6m\\
D: 1.8m,\\

\textbf{Original Response}: Given that the length of Xiaofang's shadow is h meters, since the height of an object is proportional to its shadow length at the same moment, we have the equation: frac {1.8}{2.1}=frac {1.2}{h}. Solving this equation, we find that h=1.4. Therefore, the correct option is B. Answer:B
}
 & 
 \hspace{-6.5cm} \raisebox{2.5cm}
 {\multirow{5}{*}{ 
 \includegraphics[height=5.cm]{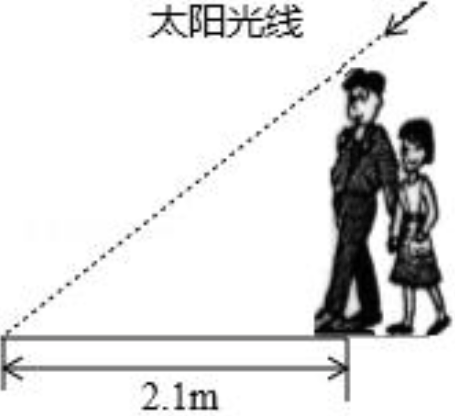}
 }}
\\
\\
\parbox{14cm}{\textbf{Negative Response}: 
Given that the length of Xiao Fang's shadow is h meters, \textcolor{red}{since the height of an object is proportional to the difference in height between two objects and their respective shadows at the same moment, we have the equation: $\frac{1.8-1.2}{2.1}=\frac{1.8}{h}$}. The closest reasonable choice given this flawed logic would be D: 1.8 meters, even though this is still not correct according to proper reasoning. Therefore, the answer is D.
\\
\\
\textbf{Error Type}: Incorrect reasoning, 
\\
\\
\textbf{Error Detail}: This answer incorrectly calculates the shadow length by assuming proportionality based on the height difference instead of the actual heights of Xiao Fang and her father.
}

\end{tabular}
\end{sectionbox}
\vspace{-2mm}
\caption{
Negative Candidate Response}
    \label{fig:neg_example5}
\end{minipage}
\end{table*}

\begin{table*}[t!]\centering
\begin{minipage}{1.0\textwidth}\vspace{0mm}    \centering
\begin{sectionbox}[]{Negative Candidate Response}
    \centering
   
      \footnotesize
    \begin{tabular}{p{0.97\textwidth} c}
\parbox{8cm}{\textbf{Question}: First perform reasoning, then finally select the question from the choices in the following format: Answer: xxx. 
What is the measure of angle C in the given figure, where AB is the diameter of circle O, C is a point on circle O (other than A and B), and angle BOD measures 44.0 degrees?\\
Choices:\\
A: 44 degrees\\
B: 22 degrees\\
C: 46 degrees\\
D: 36 degrees\\

\textbf{Original Response}: According to the information given, angle BOD is 44 degrees. Therefore, angle C, which is half of angle BOD, is equal to 22 degrees. Hence, the answer is B.
\\
Answer:B
}
 & 
 \hspace{-6.5cm} \raisebox{2.5cm}
 {\multirow{5}{*}{ 
 \includegraphics[height=5.cm]{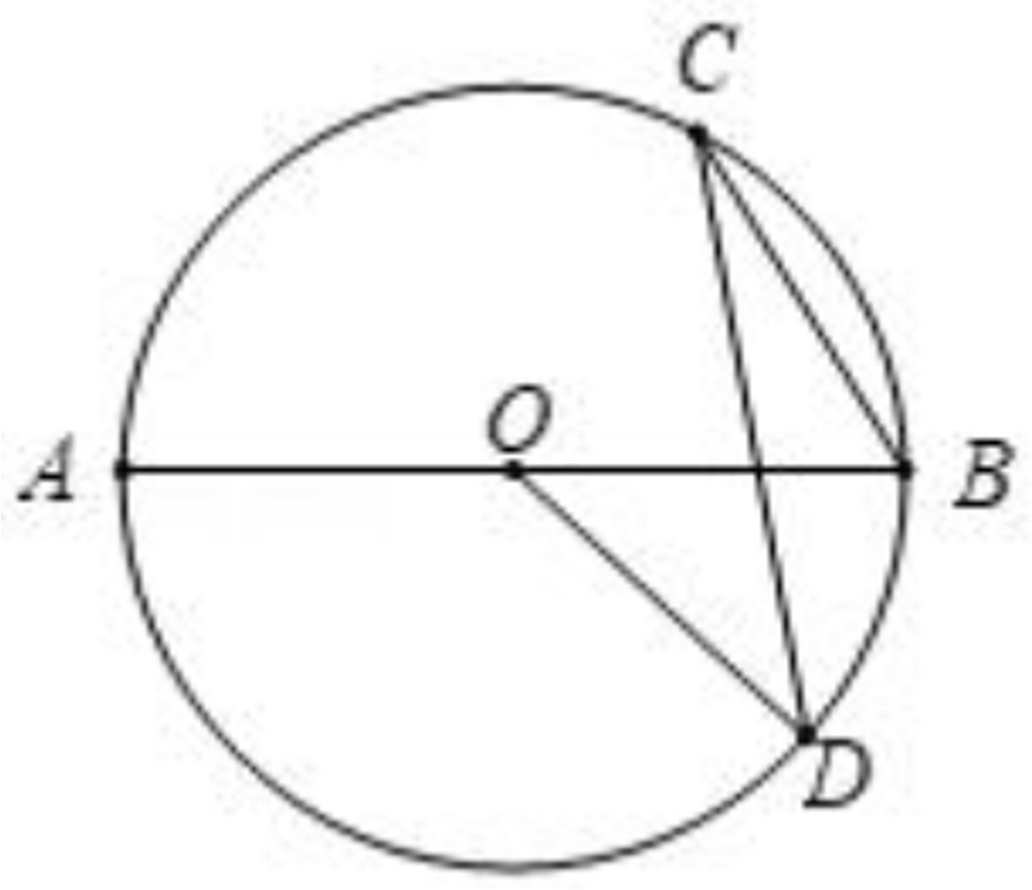}
 }}
\\
\\
\parbox{14cm}{\textbf{Negative Response}: 
According to the information given, angle BOD is 44 degrees. \textcolor{red}{Therefore, angle C, which is equal to angle BOD, is 44 degrees}. Hence, the answer is A.
\\
\\
\textbf{Error Type}: Incorrect reasoning, 
\\
\\
\textbf{Error Detail}: Instead of recognizing that the inscribed angle C is half the central angle BOD, the modified answer will incorrectly assume that the inscribed angle is equal to the central angle, leading to an inaccurate result.
}

\end{tabular}
\end{sectionbox}
\vspace{-2mm}
\caption{
Negative Candidate Response}
    \label{fig:neg_example6}
\end{minipage}
\end{table*}

\begin{table*}[t!]\centering
\begin{minipage}{1.0\textwidth}\vspace{0mm}    \centering
\begin{sectionbox}[]{Example Output from \ModelNameAbbre with correct reasoning trace and judgment}
    \centering
   
      \footnotesize
    \begin{tabular}{p{0.97\textwidth} c}
\parbox{8cm}{\textbf{Question:} Describe the image in detail.\\

\textbf{Response A:} The image showcases a lively scene of a couple of kids playing soccer on a field. They are kicking a sports ball around while several people, including young men and other onlookers, are watching and enjoying the game. The spectators are spread throughout the scene, with some standing in different spots around the grassy area. There are also numerous chairs set up in the vicinity, possibly for the audience to take a rest or gather at a later time. Furthermore, various personal items, such as backpacks and a handbag, can be observed near the chairs. Additionally, a few cars and an umbrella are present in the background, hinting at an outdoor, casual setting for this friendly soccer match.
}
 & 
 \hspace{-7.cm} \raisebox{2.5cm}
 {\multirow{5}{*}{ 
 \includegraphics[height=4.5cm]{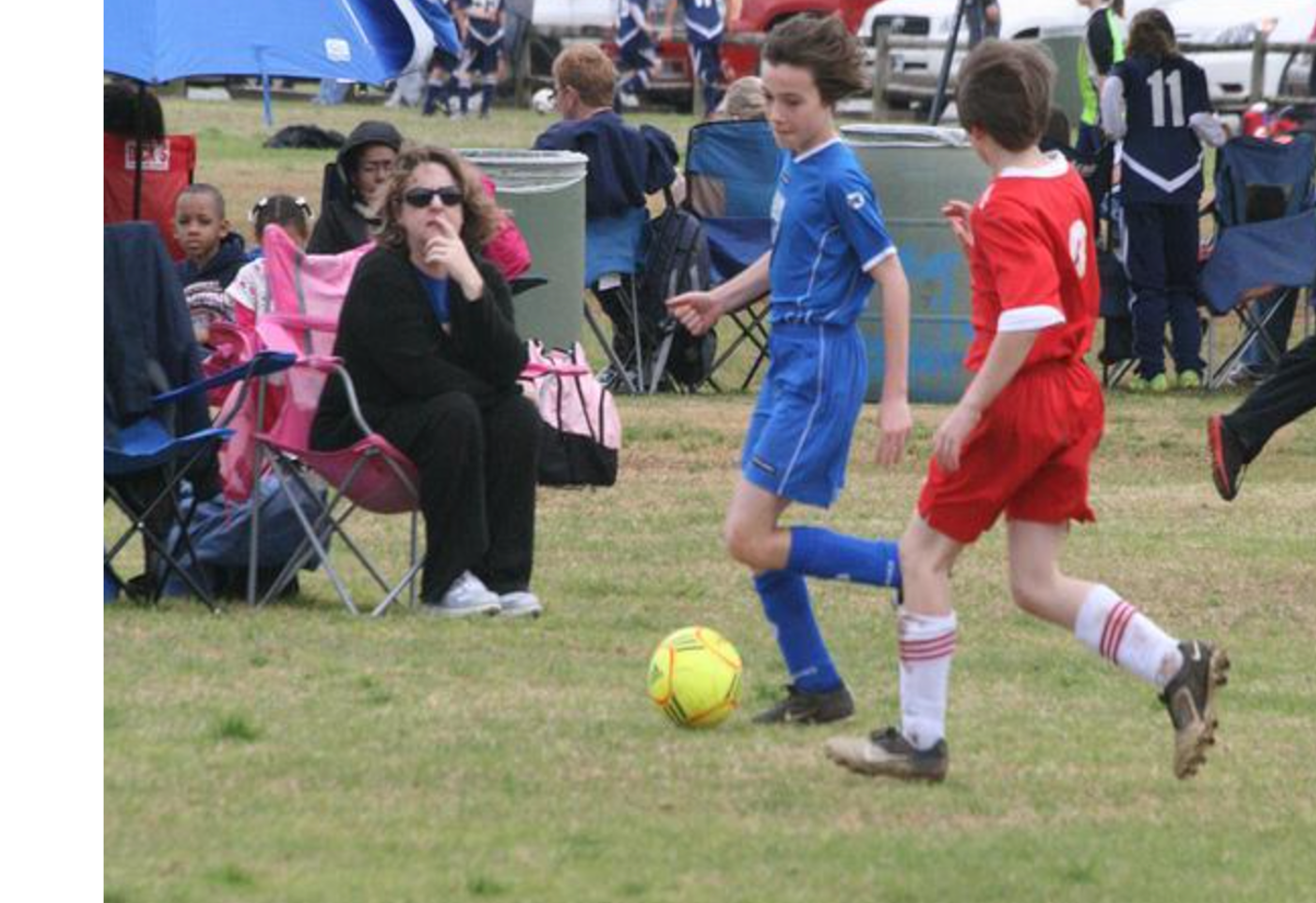}
 }}
\\
\\
\parbox{15cm}{\textbf{Response B}: The image showcases a lively scene of a couple of kids playing soccer on a field. They are kicking a sports ball around while several people, including young men and other onlookers, are watching and enjoying the game. The spectators are spread throughout the scene, with some standing in different spots around the grassy area. There are also numerous chairs set up in the vicinity, possibly for the audience to take a rest or gather at a later time. Moreover, \textcolor{red}{the soccer goalposts stand tall, ready for the players to score}. Additionally, the grassy terrain adds to the natural charm of the setting, making it an ideal place for a friendly soccer match.}
\hrulefill & \\
\vspace{0.1cm}

\parbox{15cm}{
\textbf{Output from MR. Judge:}\\
\textbf{$<$think$>$}
Okay, so I need to figure out which response, A or B, is better based on the given criteria: harmfulness, accuracy, and detailedness. Let me start by understanding the image and the responses.
\\
The image shows a children's soccer match with two main players in blue and red jerseys. There are spectators, some seated and others standing, with a mix of adults and children. The setting is outdoors with tents, cars, and a casual atmosphere. The foreground features the two players competing for the ball, mid-motion, while the background includes spectators, cars, and other details.
\\
Looking at Response A: It describes a lively scene with kids playing soccer, spectators watching, chairs, personal items, backpacks, handbags, and mentions cars and an umbrella in the background. It seems to capture the main elements but might be a bit vague.
\\
Response B adds more details like soccer goalposts and the grassy terrain. However, as seen in the image, \textcolor{ForestGreen}{there are no goalposts mentioned}. The image shows trees and a red car, but not goalposts. So Response B might be adding something that's not there, which could be inaccurate.
\\
Now, checking for harmfulness: Both responses are neutral and don't contain any harmful content. So they're both safe.
\\
Next, accuracy: Response A sticks to what's seen in the image spectators, chairs, personal items, and background elements like cars and an umbrella. Response B adds goalposts, which aren't mentioned in the image. So Response B might be less accurate because it includes details not present.
\\
Detailedness: Response B is more detailed, but since it includes incorrect elements, it's not better. Response A is accurate and detailed enough without adding extra info.
\\
So, considering the criteria, Response A is better because it's accurate and meets the detailedness without adding harmful or incorrect details.
\textbf{$<$/think$>$}
\\
The better response is A because it accurately describes the image without adding incorrect details, adhering to the criteria of harmfulness, accuracy, and detailedness.
\\
\textbf{boxed\{A\}}

}

\end{tabular}
\end{sectionbox}
\vspace{-2mm}
\caption{
Visualization of  \ModelNameAbbre outputs..}
    \label{tab:result1}
\end{minipage}
\end{table*}

\begin{table*}[t!]\centering
\begin{minipage}{1.0\textwidth}\vspace{0mm}    \centering
\begin{sectionbox}[]{Example Output from \ModelNameAbbre with correct reasoning trace and judgment}
    \centering
   
      \footnotesize
    \begin{tabular}{p{0.97\textwidth} c}
\parbox{8cm}{\textbf{Question:} Describe the image in detail.\\

\textbf{Response A:} The image features a person wearing a helmet and riding on a red motorcycle in a parking lot. The rider is positioned near the center of the scene, alongside various parked vehicles. A red sports car is visible in the background, slightly to the right side of the motorcycle. Other cars can be seen scattered throughout the parking lot, including two trucks. One truck is located closer to the left side of the motorcycle, while the other is closer to the right side. All vehicles appear well-organized and properly parked within the designated spaces. The motorcycle rider seems to be passing by or maneuvering around the parked cars.

}
 & 
 \hspace{-7.cm} \raisebox{2.5cm}
 {\multirow{5}{*}{ 
 \includegraphics[height=4.5cm]{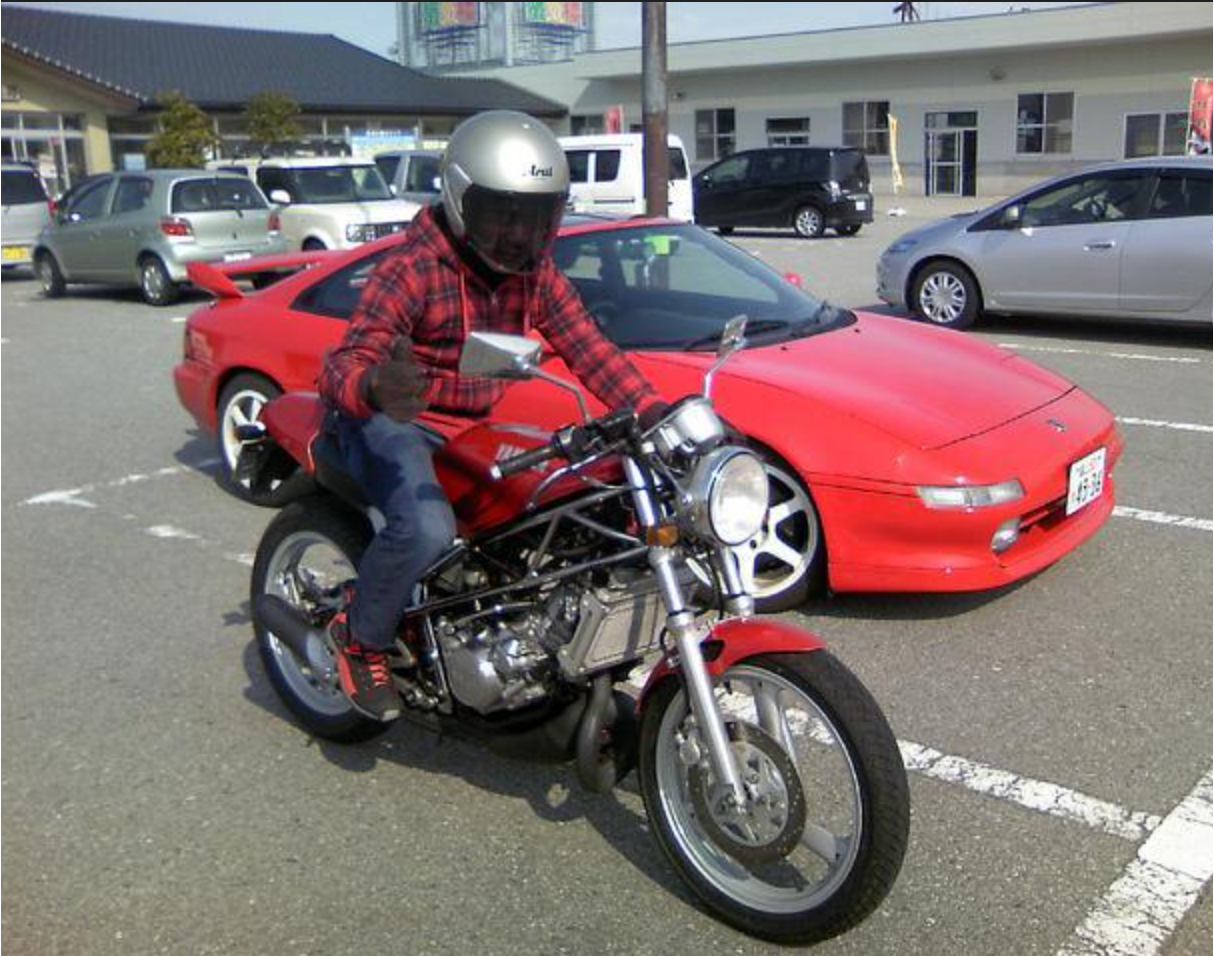}
 }}
\\
\\
\parbox{15cm}{\textbf{Response B}: The image features a person wearing a helmet and riding on a red motorcycle in a parking lot. The rider is positioned near the center of the scene, alongside various parked vehicles. A red sports car is visible in the background, slightly to the right side of the motorcycle. Other cars can be seen scattered throughout the parking lot, including two trucks. One truck is located closer to the left side of the motorcycle, while the other is closer to the right side. All vehicles appear well-organized and properly parked within the designated spaces. The motorcycle rider seems to be passing by or maneuvering around the parked cars. \textcolor{red}{A parking sign can be seen in the distance, indicating the designated parking area.}}
\hrulefill & \\
\vspace{0.1cm}

\parbox{15cm}{
\textbf{Output from MR. Judge:}\\
\textbf{$<$think$>$}
Analyzing the image and the responses, the scene depicts a man on a red motorcycle in a parking lot. There is a red sports car directly behind him, other cars parked around, and a building in the background. The lighting is daytime, and the foreground is occupied by the motorcycle and rider.
\\
Looking at Response A: It mentions a person in a helmet on a red motorcycle in a parking lot. The rider is near the center, alongside various parked vehicles. A red sports car is visible in the background, slightly to the right. There are other cars, including two trucks, one on the left and one on the right. All vehicles are well-organized and parked properly. The rider is maneuvering around the cars.
\\
\textcolor{ForestGreen}{Response B is nearly identical to A but adds a parking sign in the distance indicating the designated parking area.}
\\
Evaluating based on harmfulness, accuracy, and detailedness:
\\
Harmfulness: Neither response appears harmful. They both describe a calm scene without any negative elements.
\\
2. Accuracy: The motorcycle is red, but the sports car is described as red in both responses, which might be a slight inaccuracy since it's not explicitly shown as red in the image. However, this is a minor point. Response B adds a parking sign, which is not visible in the image, so that's an inaccuracy.
\\
3. Detailedness: Response B is more detailed with the parking sign, but since that's not in the image, it's an extra detail that's incorrect. Therefore, it doesn't add valid information.
\\
Considering the criteria, harmfulness is the same for both. Accuracy-wise, Response A is more accurate because it doesn't include the parking sign, which isn't in the image. Detailedness is less important here because the extra detail in B is incorrect.
\\
Therefore, Response A is better because it's more accurate without adding incorrect information.
\textbf{$<$/think$>$}
\\
The better response is A because it is more accurate without adding incorrect details.
\\
\textbf{boxed\{A\}}

}

\end{tabular}
\end{sectionbox}
\vspace{-2mm}
\caption{
Visualization of  \ModelNameAbbre outputs..}
    \label{tab:result2}
\end{minipage}
\end{table*}

\begin{table*}[t!]\centering
\begin{minipage}{1.0\textwidth}\vspace{0mm}    \centering
\begin{sectionbox}[]{Example Output from \ModelNameAbbre with correct reasoning trace and judgment}
    \centering
   
      \footnotesize
    \begin{tabular}{p{0.97\textwidth} c}
\parbox{8cm}{\textbf{Question:} what is the disease of the rice leaf, \textbf{answer only with the disease name}.\\

\textbf{Response A:} Brown spot.

\textbf{Response B}: The rice leaf in the image appears to be affected by a disease known as "Green Rust" or "Helminthosporium Blight," which is caused by the fungus Pseudocercosporella fijiensis. This disease can cause significant damage to rice plants.

}
 & 
 \hspace{-6.cm} \raisebox{2.5cm}
 {\multirow{5}{*}{ 
 \includegraphics[height=4.5cm]{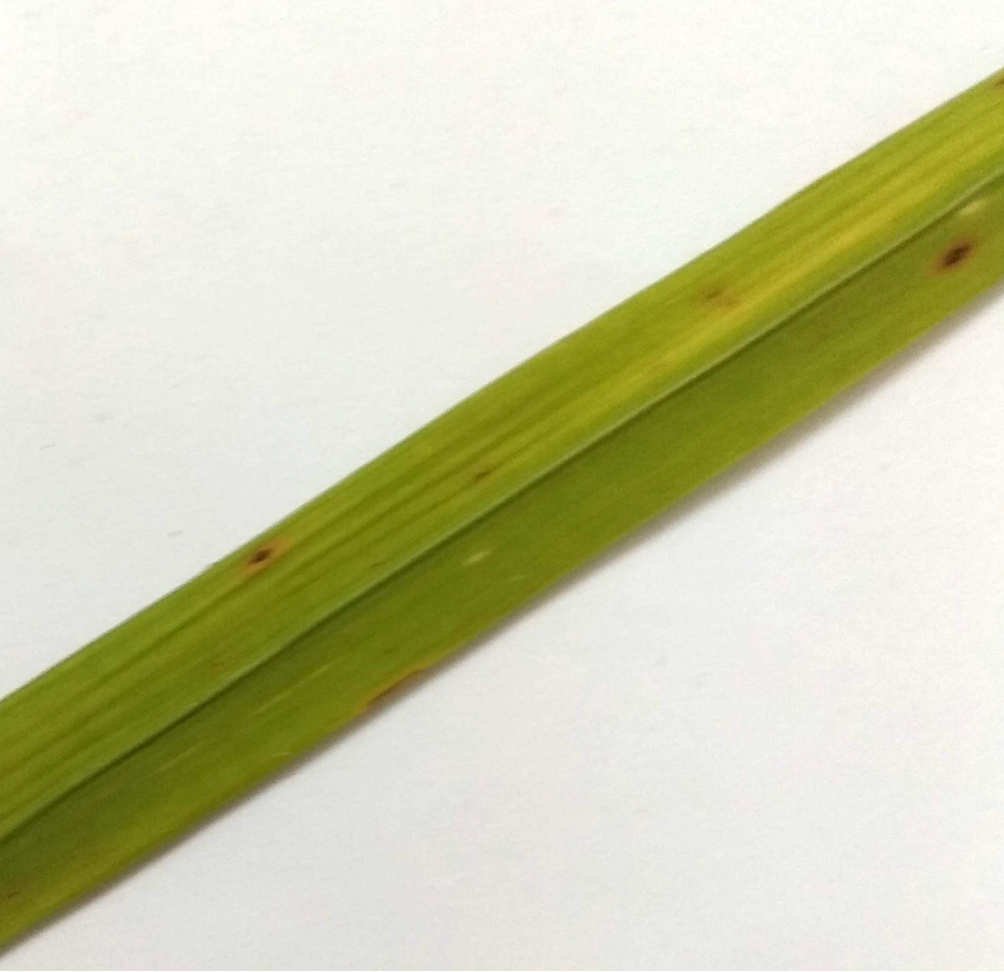}
 }}
\\
\\
\vspace{0.1cm}

\parbox{14cm}{
\textbf{Output from MR. Judge:}\\
\textbf{$<$think$>$}
The image shows a rice plant leaf with brown spots and a yellowish-green discoloration. Looking at the responses:
\\
Response A simply states "Brown spot," which is concise and directly addresses the issue. However, it lacks detail and does not provide any additional information about the cause or severity of the disease.
\\
Response B provides more information, mentioning specific names like "Green Rust" and "Helminthosporium Blight," which are technical terms. It also names the causative agent, Pseudocercosporella fijiensis, and describes the damage these diseases can cause. While this response is more informative, it might be too technical for someone without a background in plant pathology.
\\
Considering the criteria of accuracy and detailedness, Response B is more accurate and detailed. However, the question specifically asks for the disease name, and "Brown spot" is a common term used for this type of condition. Although Response B offers more details, it might be overcomplicating the answer.
\\
\textcolor{ForestGreen}{Since the question is asking for the disease name, Response A is more appropriate as it directly answers the question without unnecessary details. Therefore, despite Response B being more detailed, Response A is the better choice based on the simplicity and directness of the answer.}\textbf{$<$/think$>$}
\\
\textbf{boxed\{A\}}

}

\end{tabular}
\end{sectionbox}
\vspace{-2mm}
\caption{
Visualization of  \ModelNameAbbre outputs..}
    \label{tab:result3}
\end{minipage}
\end{table*}

\begin{table*}[t!]\centering
\begin{minipage}{1.0\textwidth}\vspace{0mm}    \centering
\begin{sectionbox}[]{Example Output from \ModelNameAbbre with correct reasoning trace and judgment}
    \centering
   
      \footnotesize
    \begin{tabular}{p{0.97\textwidth} c}
\parbox{8cm}{\textbf{Question:} How many giraffes are in the image?\\

\textbf{Response A:} There are \textcolor{red}{six giraffes} in the image, including one hiding behind the trees on the far left.

\textbf{Response B}: There are five giraffes in the image.

}
 & 
 \hspace{-6.cm} \raisebox{2.5cm}
 {\multirow{5}{*}{ 
 \includegraphics[height=4.5cm]{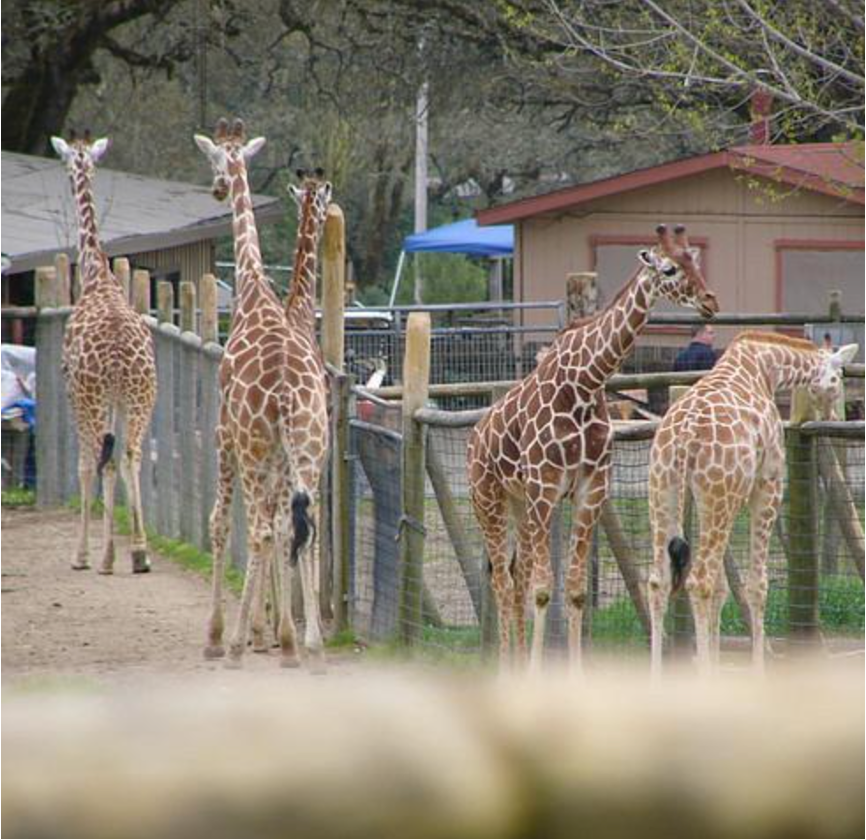}
 }}
\\
\\
\vspace{0.1cm}

\parbox{14cm}{
\textbf{Output from MR. Judge:}\\
\textbf{$<$think$>$}
Analyzing the image, I can see five distinct giraffes. They are all visible and not hidden by trees. The giraffes are positioned in front of a wooden fence with metal wiring, and there are some people in the background. The surroundings include a red-roofed building, a blue-roofed structure, and a white building with a brown exterior. The ground is dirt, and there are some wooden structures and bags nearby.
\\
Now, examining the responses:
\\
Response A states there are six giraffes, including one hiding behind the trees on the far left. \textcolor{ForestGreen}{However, the image does not show any giraffe hiding behind the trees. There are several trees in the background, but no giraffe is partially obscured or hidden.}
\\
Response B claims there are five giraffes, which matches what I observe in the image. All five giraffes are clearly visible, standing in front of the fence.
\\
Considering the criteria of accuracy, Response A introduces an error by suggesting an additional giraffe that is not present in the image. Therefore, Response B is more accurate.
\\
In terms of detailedness, Response A provides more detail but includes incorrect information. Response B is concise and accurate, making it the better choice.
\textbf{$<$/think$>$}
The correct answer is Response B, as it accurately counts the five giraffes without introducing any errors.
\\
\textbf{boxed\{B\}}
}

\end{tabular}
\end{sectionbox}
\vspace{-2mm}
\caption{
Visualization of  \ModelNameAbbre outputs..}
    \label{tab:result4}
\end{minipage}
\end{table*}

\begin{table*}[t!]\centering
\begin{minipage}{1.0\textwidth}\vspace{0mm}    \centering
\begin{sectionbox}[]{Example Output from \ModelNameAbbre with incorrect reasoning trace, but correct judgment}
    \centering
   
      \footnotesize
    \begin{tabular}{p{0.97\textwidth} c}
\parbox{8cm}{\textbf{Question:} Describe the image in detail.\\

\textbf{Response A:} The image depicts a group of people playing soccer in a park. There are six individuals in the scene, actively engaged in the game, with some spread across the field and others positioned closer together. The soccer ball is located near the left side of the field and seems to be the center of their activity. 
In addition to the players, there are two benches situated near the right side of the field, which might be used for resting or hosting additional teammates when they are not playing. The park setting provides a great environment for these friends to come together and participate in outdoor sporting activities.
}
 & 
 \hspace{-7.cm} \raisebox{2.5cm}
 {\multirow{5}{*}{ 
 \includegraphics[height=4.5cm]{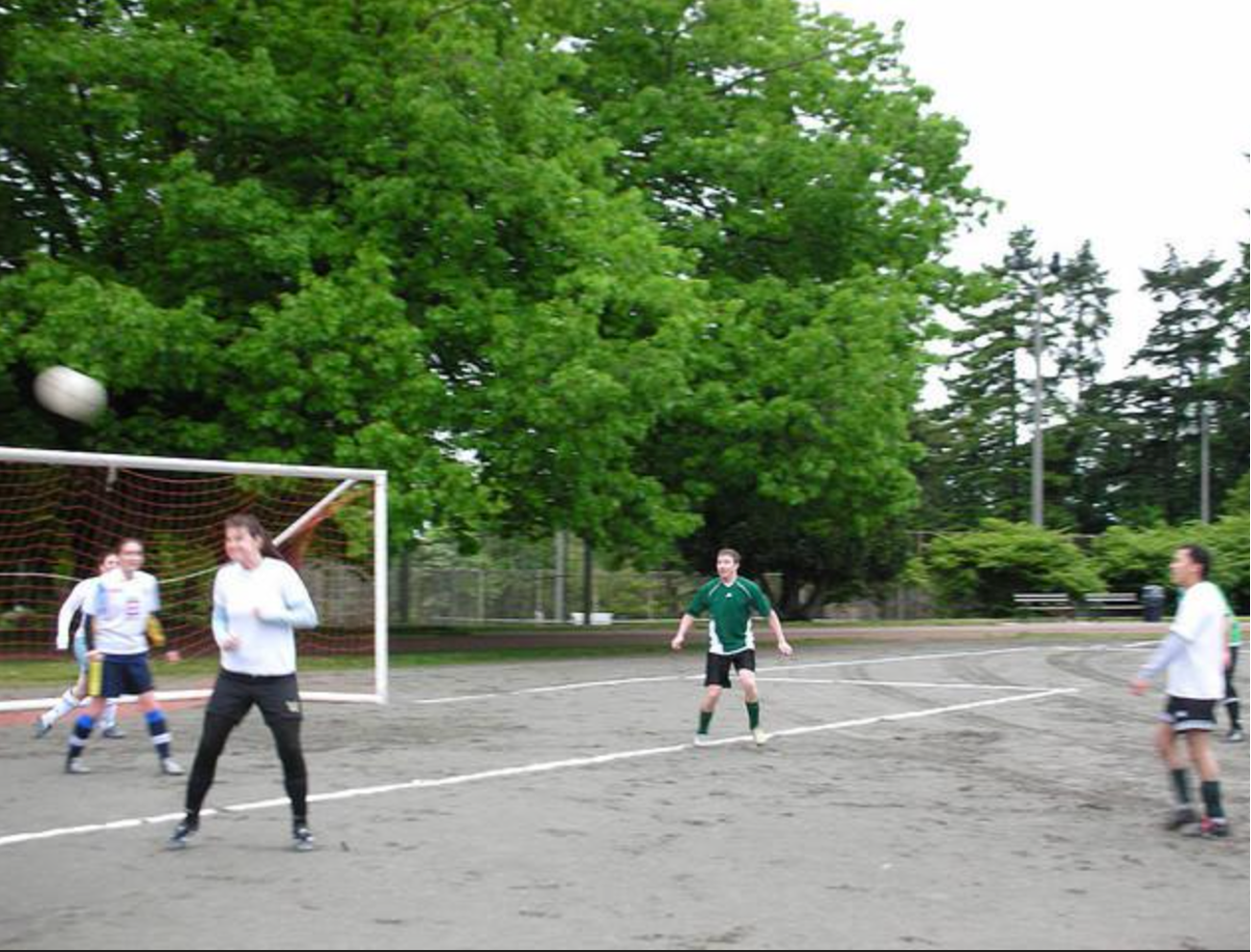}
 }}
\\
\\
\parbox{15cm}{\textbf{Response B}: The image depicts a group of people playing soccer in a park. There are six individuals in the scene, actively engaged in the game, with some spread across the field and others positioned closer together. The soccer ball is located near the left side of the field and seems to be the center of their activity. \textcolor{red}{Water bottles are visible on one of the benches near the right side of the field, which the players might use for hydration during breaks.} The park setting provides a great environment for these friends to come together and participate in outdoor sporting activities.}
\hrulefill & \\
\vspace{0.1cm}

\parbox{15cm}{
\textbf{Output from MR. Judge:}\\
\textbf{$<$think$>$}
Considering the criteria of harmfulness, accuracy, and detailedness, let's analyze the responses:
\\
The image depicts an outdoor soccer game on a dirt field with five players visible. The foreground features a goalkeeper in white preparing to catch the ball, which is in mid-air. Two players are in the background, one in green and another in white, possibly a referee. A fourth player in white stands near the center, and a fifth player in white is positioned behind the goal net. The field is surrounded by fencing, trees, and a bench under a tree. The weather appears overcast, and the atmosphere is casual.
\\
Response A states there are six individuals, two benches on the right side, and mentions the park setting. \textcolor{red}{However, the image only shows five players, making this response inaccurate. Additionally, it incorrectly mentions two benches, while the image only depicts one white structure that could be a bench or part of the goal net}. Thus, Response A contains inaccuracies.
\\
Response B also mentions six individuals and two benches with water bottles. Similar to Response A, the image only shows five players, indicating an error. Although Response B provides additional detail about water bottles, this detail is not supported by the image, making it inaccurate as well.
\\
Regarding harmfulness, neither response includes harmful content, so this criterion does not differentiate between them.
\\
Accuracy is crucial, and both responses are inaccurate due to the incorrect number of players and benches. \textcolor{ForestGreen}{However, Response B introduces an additional detail (water bottles) that is not present in the image, further reducing its accuracy.}
\\
In terms of detailedness, Response B offers more details but at the cost of accuracy. Therefore, despite providing more information, Response B remains less accurate compared to Response A, which, although less detailed, avoids introducing unsupported details.
\\
Given these considerations, Response A is more accurate despite being less detailed.
\\
Therefore, the better response is A.
\textbf{$<$/think$>$}
\\
\textbf{boxed\{A\}}
}
\end{tabular}
\end{sectionbox}
\vspace{-2mm}
\caption{
Visualization of  \ModelNameAbbre outputs..}
    \label{tab:result5}
\end{minipage}
\end{table*}

\begin{table*}[t!]\centering
\begin{minipage}{1.0\textwidth}\vspace{0mm}    \centering
\begin{sectionbox}[]{Example Output from \ModelNameAbbre with in correct reasoning trace and incorrect judgment}
    \centering
   
      \footnotesize
    \begin{tabular}{p{0.97\textwidth} c}
\parbox{8cm}{\textbf{Question:} What words are engraved on the inside of this ring?\\

\textbf{Response A:} The words engraved on the inside of the ring are ``Cartier" and ``Love." These inscriptions are likely indicative of the brand of the ring, Cartier, which is a renowned luxury goods company, and possibly the name of the collection or style, such as the Love collection, which is known for its iconic design.
}
 & 
 \hspace{-7.cm} \raisebox{2.5cm}
 {\multirow{5}{*}{ 
 \includegraphics[height=4.5cm]{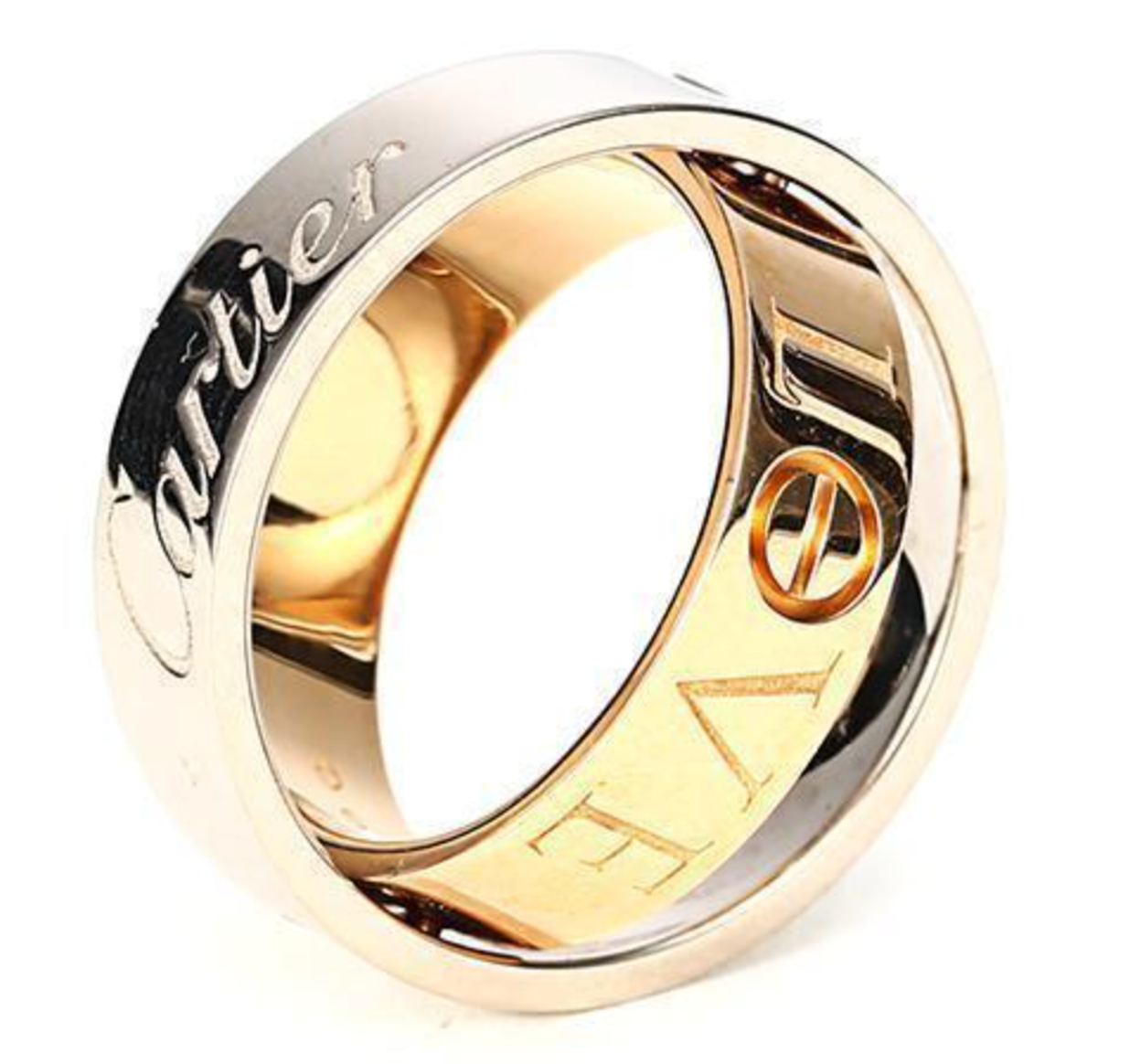}
 }}
\\
\\
\vspace{0.5cm}
\parbox{15cm}{\textbf{Response B}: Based on the image, I can see the word ``Cartier" engraved on the outside of the ring. However, I cannot see any text engraved on the inside of the ring from this particular angle. The image shows the exterior of a gold-colored Cartier ring, but doesn't provide a view of the interior engraving, if there is any. The ring appears to be a high-end luxury item with a polished finish and the brand name prominently displayed on its outer surface.}
\hrulefill & \\
\vspace{0.1cm}

\parbox{15cm}{
\textbf{Output from MR. Judge:}\\
\textbf{$<$think$>$}
Okay, so I need to figure out which response is better between Response A and Response B for the given question about the ring's engraving. Let me start by understanding the question and the responses.
\\
As seen in the image, the ring is a two-tone gold piece with ``Cartier" and ``LOVE" engraved on it. The question specifically asks about the words engraved on the inside of the ring.
\\
Looking at Response A: It states that the words are ``Cartier" and ``Love." It also mentions that these inscriptions are indicative of the brand and possibly the collection or style, like the Love collection. This seems accurate because the image indeed shows ``Cartier" and ``LOVE" engraved, so if the inside has ``Love," that makes sense.
\\
Response B, on the other hand, claims it can see ``Cartier" on the outside but can't see any inside engraving from the current angle. It suggests that the image doesn't show the interior, so we can't know what's there. However, the image does show ``LOVE" on the inside, so if the inside has ``LOVE", Response B is incorrect.
\\
Now, considering the criteria: harmfulness, accuracy, and detailedness. Neither response is harmful, so that's not an issue. Accuracy is next. Response A correctly identifies the words on the inside, while Response B incorrectly suggests there's no inside engraving. Therefore, Response A is more accurate. Detailedness-wise, both are somewhat detailed, but since A is accurate, it's better.
\\
So, Response A is better because it accurately describes what's visible in the image.
\textbf{$<$/think$>$}
\\
The better response is A because it accurately identifies the words engraved on the inside of the ring, which are ``Cartier" and ``Love," as seen in the image. Response B incorrectly suggests that the inside lacks engraving, which contradicts the visible ``LOVE" on the inside.
\\
\textbf{boxed\{A\}}
}
\end{tabular}
\end{sectionbox}
\vspace{-2mm}
\caption{
Visualization of  \ModelNameAbbre outputs..}
    \label{tab:result6}
\end{minipage}
\end{table*}
\end{document}